\documentclass[a4paper,9pt,twocolumn]{article}
\usepackage[utf8]{inputenc}
\usepackage{times}
\usepackage{epsfig}
\usepackage{graphicx}
\usepackage{amsmath}
\usepackage{amssymb}
\usepackage{color}
\usepackage[usenames,dvipsnames]{xcolor}
\usepackage{tabularx}
\usepackage{multirow}
\usepackage{hhline}
\usepackage{subfig}
\usepackage{booktabs}
\usepackage{epstopdf}
\usepackage{amsmath}
\usepackage{xspace}
\usepackage{stfloats}
\usepackage{authblk}

\newcommand{\Real}{\mathbb{R}}

\newcommand{\argmin}[1]{\underset{#1}{\operatorname{arg}\,\operatorname{min}}\;}

\newcommand{\name}{rank pooling\xspace}
\newcommand{\Name}{Rank pooling\xspace}
\def\etal{\emph{et al}\xspace}
\def\ie{\emph{i.e.}\xspace}
\def\eg{\emph{e.g.}\xspace}

\title{Rank Pooling for Action Recognition}

\author[1]{Basura~Fernando\thanks{basura.fernando@anu.edu.au}}
\author[2]{Efstratios~Gavves\thanks{E.Gavves@uva.nl}}
\author[3]{Jos\'{e}~Oramas~M\thanks{jose.oramas@esat.kuleuven.be}}
\author[3]{Amir~Ghodrati\thanks{amir.ghodrati@esat.kuleuven.be}}
\author[3]{Tinne~Tuytelaars\thanks{tinne.tuytelaars@esat.kuleuven.be}}
\affil[1]{ACRV, The Australian National University, Australia}
\affil[2]{QUVA Lab, University of Amsterdam, Netherlands}
\affil[3]{KU Leuven, ESAT-PSI, iMinds, Belgium}

\begin{document}

\maketitle

\begin{abstract}
We propose a function-based temporal pooling method that captures the latent structure of the video sequence data - e.g. how frame-level features evolve over time in a video. We show how the parameters of a function that has been fit to the video data can serve as a robust new video representation. As a specific example, we learn a pooling function via ranking machines. By learning to rank the frame-level features of a video in chronological order, we obtain a new representation that captures the video-wide temporal dynamics of a video, suitable for action recognition. Other than ranking functions, we explore different parametric models that could also explain the temporal changes in videos. The proposed functional pooling methods, and rank pooling in particular, is easy to interpret and implement, fast to compute and effective in recognizing a wide variety of actions. We evaluate our method on various benchmarks for generic action, fine-grained action and gesture recognition. Results show that rank pooling brings an absolute improvement of 7-10 average pooling baseline. At the same time, rank pooling is compatible with and complementary to several appearance and local motion based methods and features, such as improved trajectories and deep learning features.
\end{abstract}

\section{Introduction}
\label{sec:action-intro}
A recent statistical study has revealed more than 300 hours of video content are added to YouTube every minute~\cite{youtubestat}. 
Moreover, a recent survey on network cameras has indicated that a staggering 28 million network cameras will be sold in 2017 alone~\cite{Su2015}.
Given the steep growth in video content all over the world, the capability of modern computers to process video data and extract information from them remains a huge challenge.
As such, human action and activity recognition in realistic videos is of great relevance.

Most of the progress in the field of action recognition over the last decade 
has been associated with either of the following two developments. The first development has been the
 {\em local spatio-temporal descriptors}, including spatio-temporal~\cite{Laptev2005} and densely sampled~\cite{Laptev2008, DeGeest2014} interest points, dense trajectories~\cite{wang2013dense}, and 
motion-based gradient descriptors~\cite{jain2013better}. The second development has been the 
adoption of {\em powerful encoding schemes} with an already proven track record in 
object recognition, such as Fisher Vectors~\cite{wang2013action}. 
Despite the increased interest in action~\cite{Laptev2005,  Laptev2008, wang2013dense, jain2013better, DeGeest2014, karpathy2014large, Simonyan2014} and event~\cite{Tang2012, izadinia2012recognizing, Revaud2013, MazloomTMM14} recognition, however, relatively few works have dealt with the problem of modeling the temporal information within a video.

\begin{figure}[t!]
    \centering 
    \includegraphics[width=.9\linewidth]{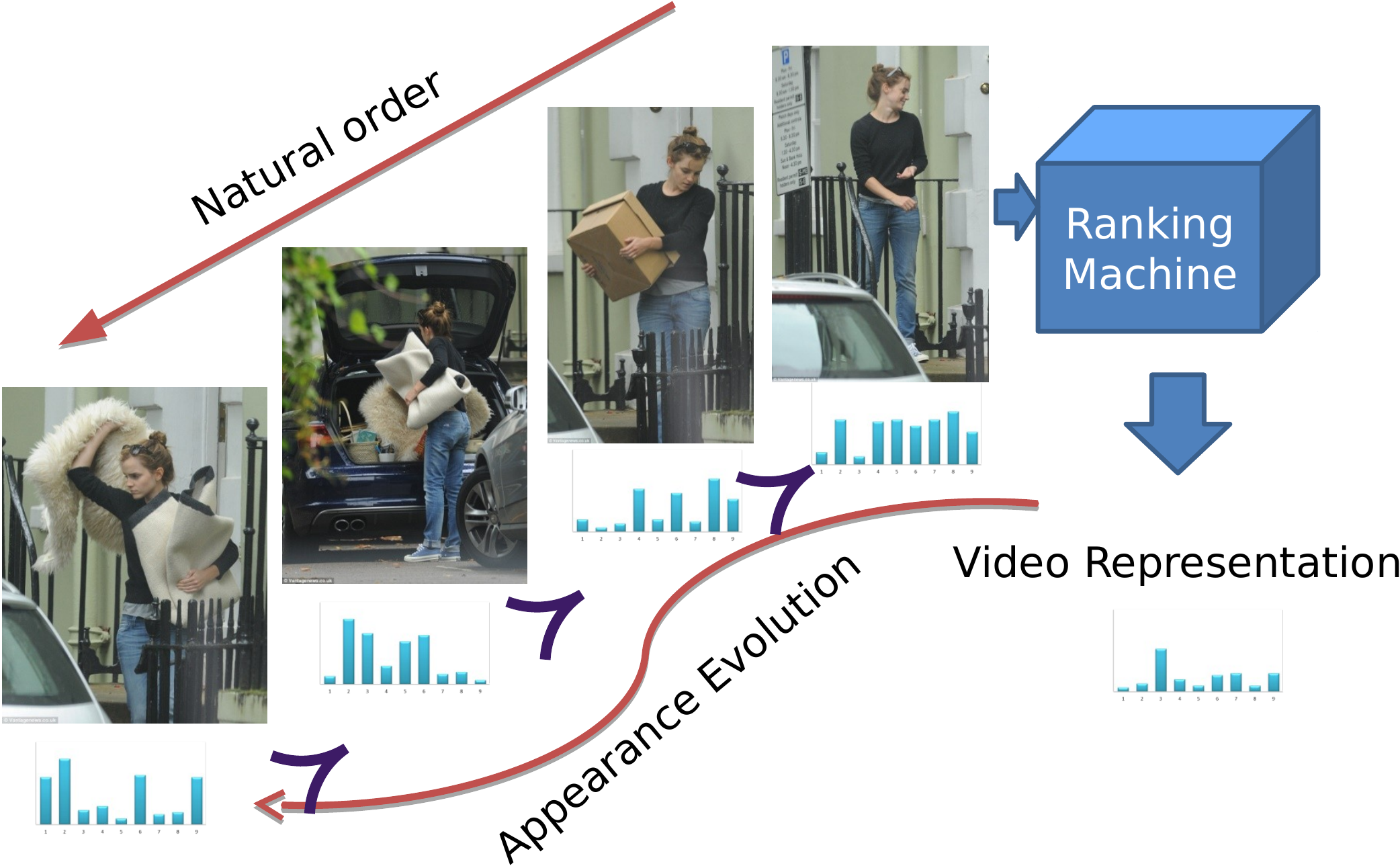}   
     \caption{Illustration of how \name works. In this video, as Emma 
\textit{moved out from the house}, the appearance of the frames 
evolves with time. A ranking machine learns 
this evolution of the appearance over time and returns a ranking function. We 
use the parameters of this ranking function as a new video representation which 
captures vital information about the action.}
    \label{fig:daewin}
\end{figure}

Modeling the \emph{video-wide temporal evolution} of appearance in videos 
is a challenging task, due to the large variability and complexity of video 
data.
Not only actions are performed at largely varying speeds for different videos, but often the speed of the 
action also varies non-linearly even within a single video.
Hence, while methods have been proposed to model the {\em video-wide temporal 
evolution} in actions (\eg using HMM~\cite{Wang2011,Wu2014}, CRF-based methods 
\cite{Song2013} or deep networks~\cite{taylor2010convolutional}), the impact of 
these on action recognition performance so far has been somewhat disappointing. 
What is more, simple but robust techniques such as temporal pyramids that are similar to spatially dividing images~\cite{Lazebnik2006} and objects~\cite{Gavves2014} are insufficient.
Nevertheless, it is clear that many actions and activities have a characteristic temporal ordering. See for instance the ``moving out of the house'' action in Figure~\ref{fig:daewin}. Intuitively, one 
would expect that a video representation that encodes this temporal change of appearances 
should help to better distinguish between different actions. Obtaining a good video-wide representation from a video still remains a challenge.

In this paper, we propose a new video representation that captures this video-wide 
temporal evolution. We start from the observation that, even if the execution 
time of actions varies greatly, the \emph{temporal ordering is typically 
preserved}. We propose to capture the temporal ordering {\em of a particular 
video} by training a linear ranking machine on the frames of  that video. More 
precisely, given all the frames of the video, we learn how to arrange them in 
chronological order, based on the content of the frames. The parameters of the 
linear ranking function encodes the video-wide temporal evolution of appearance 
of that video in a principled way. To learn such ranking machines, we use the 
supervised learning to rank framework~\cite{Liu2009}. Ranking machines trained 
on different videos of the same action can be expected to have similar 
ranking functions. Therefore, we propose to use the parameters of the 
ranking machine as a new video representation for action recognition. 
Classifiers trained on this new representation turn out to be remarkably good 
at distinguishing actions. Since the ranking machines act on frame content, they actually capture both 
the appearance and their evolution over time. We call our method \textbf{\emph{\name}}.

The key contribution of \name is to use the parameters of the ranking functions as a new 
video representation that captures the \textit{video-wide temporal evolution of 
the video}. Our new video representation is based on a principled learning 
approach, it is efficient and easy to implement. Last but not least, with the 
new representation we obtain state-of-the art results in action and gesture 
recognition. The proposed use of parameters of functions as a new representation is by no means restricted
to action recognition or ranking functions. Other than ranking functions, we explore
different parametric models, resulting in a whole family of pooling operations.

The rest of the paper is organized as follows: in 
Section~\ref{sec:action-related} we position our work w.r.t. existing work. 
Sections~\ref{sec:action-theory} and \ref{sec:action-videorepr} describe our method, 
while Section~\ref{sec:action-overview} provides an insight from its application 
for action classification.
This is followed by the evaluation of our method in Section~\ref{sec:action-exp}. 
We conclude this paper 
in Section  \ref{sec:action-conclusion}.


\section{Related work}
\label{sec:action-related}
\subsection{Action recognition}

Capturing temporal information of videos for action recognition has been a well studied research domain. Significant improvements have been witnessed in modeling local motion patterns present in short frame sequences~\cite{Laptev2005,wang2013dense,wang2013action}. Jain \etal.~\cite{Jain_2014_CVPR} proposed to first localize the actions in the video and exploit them for refining recognition.

To avoid using hand-engineered features, deep learning methodologies~\cite{Le2011, taylor2010convolutional} have also been investigated. Dynamics in deep networks can be captured either by extending the connectivity of the network architecture in time~\cite{karpathy2014large} or by using stacked optical flow instead of frames as input for the network~\cite{Simonyan2014}. The two stream stacked convolutional independent subspace analysis method, referred to as ConvISA~\cite{Lan2015}, is a neural network architecture that learns both visual appearance and motion information in an unsupervised fashion on video volumes. A human pose driven CNN feature extraction pipeline is presented in~\cite{Cheron2015}. In~\cite{Cheron2015}, authors represent body regions of humans with motion-based and appearance-based CNN descriptors. Such descriptors are extracted at each frame and then aggregated over time to form a video descriptor. To capture temporal information, authors consider temporal differences of frames and then concatenate the difference of vectors. Two convolutional neural networks are used to capture both appearance-based and motion-based features in \emph{action tubes}~\cite{Gkioxari2015}. In this method~\cite{Gkioxari2015}, the first spatial-CNN network takes RGB frames as input and captures the appearance of the actor as well as other visual cues from the scene. The second network, referred as the motion-CNN, operates on the optical flow signal and captures the movement of the actor. Spatio-temporal features are extracted by combining the output from the intermediate layers of the two networks. The benefits of having objects in the video representation for action classification is presented in~\cite{Jaincvpr15}.

Although the aforementioned methods successfully capture the local changes within small time windows, they are not designed to model the higher level motion patterns and video-wide appearance and motion evolution associated with certain actions.

\subsection{Temporal and sequential modeling}

State-space models such as generative Hidden Markov Models (HMMs) or discriminative Conditional Random Fields (CRFs) have been proposed to model dynamics of videos since the early days~\cite{Yamato1992,Sminchisescu2006}. Generative methods such as HMMs usually learn a joint distribution over both observations and action labels. In these early works, most often, the observations consist of visual appearance or local motion feature vectors obtained from videos. This results in HMMs that learn the appearance or the motion evolution of a specific action class. Then the challenge is to learn all variations of a single action class. Given the complexity, variability and the subtle differences between action classes, these methods may require a lot of training samples to learn meaningful joint probability distributions.

Discriminative CRF methods learn to discriminate two action classes by modeling conditional distribution over class labels. However, similar to HMMs, CRFs may also require a large amount of training samples to estimate all parameters of the models. In contrast, our proposed method does not rely on class labels to encapsulate temporal information of a video sequence. The proposed method captures video specific dynamic information and relies on standard discriminative methods such as SVM to discriminate action classes.

More recently, new machine learning approaches based on CRF, HMM and action grammars, have been researched for action recognition~\cite{raptis2012discovering, ryoo2006recognition, Song2013, Tang2012, Wang2011} by modeling higher level motion patterns. In~\cite{Wang2011}, a part-based approach is combined with large-scale template features to obtain a discriminative model based on max-margin hidden conditional random fields. In~\cite{Song2013}, Song \etal. rely on a series of complex heuristics and define a feature function for the proposed CRF model. In~\cite{Tang2012} Tang \etal. propose a max-margin method 
for modeling the temporal structure in a video. They use a HMM model to capture the transitions of action appearances and duration of actions. 

Temporal ordering models have also been applied in the context of complex activity recognition~\cite{izadinia2012recognizing, Ponce2015, rohrbach2012script, Sun2013}. They mainly focus on inferring composite activities from pre-defined, semantically meaningful, basic-level action detectors. In~\cite{Sun2013}, a representation for events is presented that encodes statistical information of the atomic action transition probabilities using a HMM model. In \cite{Ponce2015}, a set of shared spatio-temporal primitives, subgestures, are detected using genetic algorithms. Then, the dynamics of the actions of interest are modeled using the detected primitives and either HMMs or Dynamic Time Warping (DTW). Similar to the above works, we exploit the temporal structure of videos but in contrast, we rely on ranking functions to capture the evolution of appearance or local motion. Using the learning-to-rank paradigm, we learn a functional representation for each video.

Due to the large variability of motion patterns in a video, usually latent sequential models are not efficient. To cope with this problem, representations in the form of temporal pyramids ~\cite{gaidon2012recognizing, Laptev2008} or sequences of histograms of visual features~\cite{gaidon2011actom} are introduced. A method that aims at comparing two sequences of frames in the frequency domain using fast Fourier analysis called circulant temporal aggregation is presented in \cite{Revaud2013} for event retrieval. 
Different from the above, we explicitly model video-wide, video level dynamics using a principled learning paradigm. Moreover, contrary to~\cite{gaidon2011actom}, our representation does not require manually annotated atomic action units during training. 

Recurrent neural networks have also been extensively studied in the context of sequence generation and sequence classification~\cite{Hochreiter1997,Sutskever2014}. In~\cite{Srivastava2015} the state of the LSTM encoder after observing the last input frame is used as a video representation~\cite{Srivastava2015}. A hierarchical recurrent neural network for skeleton based action recognition is presented in~\cite{Du2015}. An LSTM model that uses CNN features for action recognition is presented in~\cite{Ng2015}. Typically, recurrent neural networks are trained in a probabilistic manner to maximize the likelihood of generating the next element of the sequence. They are conditional loglinear models. In contrast, the proposed \name uses a support vector based approach to model the elements in the sequence. \Name uses empirical risk minimization to model the evolution of the sequence data. Furthermore, in comparison to RNN-LSTM-based methods, \Name is efficient both during training and testing, and effective even for high dimensional input data. 

\subsection{Functional representations}

Our work has some conceptual similarity to the functional representations used in geometric modeling~\cite{pasko1995function}, which are used for solid and volume modeling in computer graphics. In this case an object is considered as a point set in a multidimensional space, and is defined by a single continuous real-valued function of point coordinates of the nature $f(x_1,x_2, ..., x_n)$ which is evaluated at the given point by a procedure traversing a tree structure with primitives in the leaves and operations in the nodes of the tree. The points with $f(x_1,x_2, ..., x_n) \ge 0$ belong to the object, and the points with $f(x_1,x_2, ..., x_n) < 0$ are outside of the object. The point set with $f(x_1,x_2, ..., x_n)=0$ is called an isosurface. Similarly, in our approach the ranking function has to satisfy chronological order constraints on frame feature vectors and we use the ranking function as a representation of that video.

Since we use the parameters of a linear function as a new representation of a particular sequence, our work also bears some similarity to the exemplar SVM concept~\cite{Malisiewicz2011,Zepeda_2015_CVPR}. Differently, our objective is to learn a 
representation for the relative ordering of a set of frames in a video. At the same time we do not need to rely on negative data to learn the representation, as is the case for exemplar SVM.

Meta-representation has the ability to represent a higher-order representation with a lower-order representation embedding. It is the capacity to represent a representation. Our \name representation can also be considered as a meta-representation. The parameters of the ranking function in fact represent a lower dimensional embedding of chronological structure of the frames. In the learning to rank paradigm, these ranking functions are trained to order data. Our hypothesis is that this parametric embedding of sequence data can be used to represent their dynamics. \\

This paper extends the work of~\cite{Fernando2015a}. Compared to the conference version, this paper gives a more precise account of the internals of \name. First, we provide an extended discussion of related work, covering better the recent literature. From a technical point of view, we generalize the concept of rank pooling to a framework that uses functional parameters as a new video representation. We hypothesize that any stable and robust parametric functional mapping that maps frame data to the time variable can be used for modeling the video dynamics. Furthermore, we analyze the types of non-linear kernels that best capture video evolution. We provide some empirical evidence to demonstrate the capabilities of \name.  Finally, we combine \name with convolutional neural network features to further boost the state-of-the-art action recognition performance. 

Recently, Fernando~\etal extend \name to encode higher order dynamics of a video sequence in a hierarchical manner in~\cite{Fernando2016a} and in~\cite{Bilen2016} Bilen~\etal introduced dynamic image networks which allows us to learn dynamic representation using CNNs and \name. An end-to-end CNN video classification network with rank-pooling and bi-level optimization is presented in ~\cite{FernandoICML2016}.

\section{Video representations}
\label{sec:action-theory}
In this section we present our \emph{temporal pooling} method, which encodes dynamics of video sequences and, more specifically captures the video-wide temporal evolution (VTE) of the appearance in videos. First, we present the main idea in Section~\ref{sec:vectorfields} where we propose to use parameters of suitable functions to encode the dynamics of a sequence. Then, in Section~\ref{sec:rankpooling} we present how to formulate these specific functions using rankers. Next, in Section~\ref{sec:generalization-capacity} we analyse the generalization capacity of the proposed rank pooling. Finally, in Section~\ref{sec:parameter-pooling} we describe how to use functional parameters of other parametric models as a temporal representation and compare traditional temporal pooling methods with rank pooling.

\subsection{Functional parameters as temporal representations}
\label{sec:vectorfields}

We assume that each frame of a given video is represented by a vector $\mathbf{x}$.  Then the video composed of $n$ frames is a sequence of vectors, $X=[\mathbf{x_1},\mathbf{x_2},\dots,\mathbf{x_n}]$ . A frame at discrete time step $t$ is denoted by a vector $\mathbf{x_t} \in \Real^D$. Given this sequence of vectors, we first smooth the sequence $X$ to a more general form to obtain a new sequence $V=[\mathbf{v_1},\mathbf{v_2},\dots,\mathbf{v_n}]$. We discuss how to obtain smoothed sequences in Section~\ref{sec:action-videorepr}. For the rest of the analysis we use smoothed sequences $V$, unless otherwise specified. Last, we use the notation $\mathbf{x}_{1:t}$ or $\mathbf{v}_{1:t}$ to denote a sub-sequence from time step $1$ to $t$.

Our goal is to encode the temporal evolution of appearances, or, in other words the dynamics $\mathcal{D}$ of the sequence $V$. At an abstract level, dynamics $\mathcal{D}$ reflect the way the vector valued input changes from time $t$ to $t+1$ for all $t$. Assuming that the sequence $V$ is sufficiently smooth, we can encode the dynamics of $V$ using a linear function $\Psi_u=\Psi(V; \mathbf{u})$ parametrized by $\mathbf{u}$, such that $\Psi$ approximates $\mathcal{D}$, namely
\begin{equation}
\argmin{u} ||\mathcal{D} - \Psi_u||.
\label{eq:dynamics}
\end{equation}
For a given definition of dynamics $\mathcal{D}$ (see below), there exists a family of functions $\Psi$.
Different videos from the same action category will have different (yet similar) appearances and will be characterized by different (yet similar) appearance dynamics. 
For each video $V_i(\cdot)$, we learn a different dynamics function $\Psi_i(\cdot; \mathbf{u_i})$ parametrized by $\mathbf{u_i}$. Given stability and robustness guarantees of the family of functions $\Psi$, different videos from the same action category then result in similar dynamics functions $\Psi_i(\cdot; \mathbf{u_i})$.

As the family of functions $\Psi$ for modeling the dynamics is the same for all videos, what characterizes the dynamics of each specific video is the parametrization $\mathbf{u_i}$. We propose to use the parameters $\mathbf{u_i} \in \Real^D$ of $\Psi_i$ as a new video representation, capturing the specific \textit{appearance evolution of the video}. Thus we obtain a functional representation, where the functional parameters $\mathbf{u_i}$ serve as the representation, capturing a vital part of the video-wide temporal information.

As a first concrete case, in the next section, we present how we could learn such functional representations using the learning-to-rank paradigm.

\subsection{Rank pooling}
\label{sec:rankpooling}

One way to understand dynamics $\mathcal{D}$ is to consider them as the driving force for placing frames in the correct order. Indeed, in spite of the large variability in speed, between different videos and even within a single video, the relative ordering is relatively preserved. 
To capture such dynamics for video sequence $V_i$, we consider the learning-to-rank~\cite{Liu2009} paradigm, which optimizes ranking functions of the form $\Psi(t, \mathbf{v_{1:t}; \mathbf{u}})$.
We can either employ a point-wise~\cite{Smola2004}, a pair-wise~\cite{Liu2009} or a sequence-based ranking machine~\cite{FernandoICCV15}.
Then, we can use the parameters of these ranking machines as our new video representation in a process that we coin \emph{rank pooling}.

Videos are ordered sequences of frames, where the frame order also dictates the evolution of the frame appearances. We focus on the relative orderings of the frames. If $\mathbf{v_{t+1}}$ succeeds $\mathbf{v_{t}}$ we have an ordering denoted by $\mathbf{v_{t+1}} \succ \mathbf{v_t}$. As such, we end up with order constraints $\mathbf{v_n} \succ \ldots \succ \mathbf{v_t} \succ \ldots \succ \mathbf{v_1}$. We exploit the transitivity property of video frames to formulate the objective as a pairwise learning-to-rank problem \ie (if $\mathbf{v_{a}} \succ \mathbf{v_{b}}$ and $\mathbf{v_{b}} \succ \mathbf{v_{c}} \implies \mathbf{v_{a}} \succ \mathbf{v_{c}}$).

To model the video dynamics with pair-wise rank-pooling, we solve a constrained minimization \emph{pairwise-learning-to-rank}~\cite{Liu2009} formulation, such that it satisfies the frame order constraints. Pairwise linear ranking machines learn a linear function $\psi(\mathbf{v}; \mathbf{u})= \mathbf{u}^T \cdot \mathbf{v}$ with parameters $\mathbf{u} \in \Real^D$. The ranking score of $\mathbf{v_t}$ is obtained by $\psi(\mathbf{v_t}; \mathbf{u})=\mathbf{u}^T \cdot \mathbf{v_t}$ and satisfies the pairwise constraints $(\mathbf{v_{t+1}} \succ \mathbf{v_t})$ by a large margin, while avoiding over-fitting. As a result we aim to learn a parametric vector $\mathbf{u}$ such that it satisfy all constraints $\forall t_i,t_j$, $\mathbf{v_{t_i}} \succ \mathbf{v_{t_j}} \Longleftrightarrow \mathbf{u}^T \cdot \mathbf{v_{t_i}} > \mathbf{u}^T \cdot \mathbf{v_{t_j}}$. 

Using the structural risk minimization and max-margin framework, the constrained learning-to-rank objective is
\begin{align}
\label{eq:dynamicsrank}
\argmin{\mathbf{u}} \; & \frac{1}{2}\| \mathbf{u} \|^2 + C \sum_{\forall i,j 
 \mathbf{v_{t_i}} \succ \mathbf{v_{t_j}} } \epsilon_{ij}\\ \nonumber
s.t. \; & \mathbf{u}^T \cdot (\mathbf{v_{t_i}}-\mathbf{v_{t_j}}) \ge 1 - 
\epsilon_{ij} \\ \nonumber
\; & \epsilon_{ij} \ge 0.
\end{align}
As the parameters $\mathbf{u}$ define the frame order of frames $\mathbf{v_t}$, they represent how the frames evolve with regard to the appearance of the video. Hence, the appearance evolution is encoded in the parameter $\mathbf{u}$. The above optimization objective is expressed on the basis of RankSVM~\cite{JoachimsKDD2006}, however, any other linear learning-to-rank method can be employed.
%
For example, in point-wise rank pooling we seek a direct mapping from the input time dependent vectors $v_t$ to the time variable $t$ based on the linear parameters $\mathbf{u}$. Namely, we have that
\begin{align}
\label{eq:mapping}
 g(\mathbf{v_t};\mathbf{u}) \mapsto t \\ \nonumber
 \mathbf{u^*} = \argmin{\mathbf{u}} \sum_t |t - \mathbf{u^T \cdot v_t}|.
\end{align}
The support vector regression (SVR)~\cite{Smola2004} formulation is a robust extension of equation~\ref{eq:mapping} and thus, one can use SVR parameters to encode the dynamics. Support vector regression is known to be a point-wise ranking formulation~\cite{Liu2009}. The solution of SVR would also satisfy the order constraints $g(\mathbf{v_{q}};\mathbf{u}) > g(\mathbf{v_{j}};\mathbf{u}) $ if $\mathbf{v_{q}} \succ \mathbf{v_{j}}$ because of the direct mapping of the form $g(\mathbf{v_t};\mathbf{u}) \mapsto t$.
%
%

In summary, to represent dynamics $\mathcal{D}$ of a video $V$ using rank pooling, we use the parameter vector $\mathbf{u}$ as a video representation. 
The vector $\mathbf{u}$ is a temporal encoding of the input vector sequence $\mathbf{v_n} \succ \ldots \succ \mathbf{v_t} \succ \ldots \succ \mathbf{v_1}$. 
The video representation $\mathbf{u}$ can be learnt either using a pair-wise ranking machine as in equation~\ref{eq:dynamicsrank} or using the direct mapping as in equation~\ref{eq:mapping}, \ie SVR~\cite{Smola2004} (our default setting).
Modeling the temporal evolution via rankers displays several advantages. First, in videos in the wild we typically observe a large variability in speed at which actions are performed. This is not an issue for ranker functions that are oblivious to the pace at which the frames appear and only focus on their accurate relative ordering. Second, a powerful advantage of linear ranking machines is that their function parameters reside in the same space as the input sequence data $V$.
%
\subsection{Generalization capacity}
\label{sec:generalization-capacity}

As explained above, we use the parameters of learnt ranking functions to model the temporal dynamics of the specific video.
All functions from all videos will belong to the same parametric family of models.
However, as the different videos will differ in appearance and their dynamics, each function will be characterized by a different set of parameters.
It remains to be answered whether different videos that contain the same action category will be characterized by similar parameters or not.

For action recognition we make the basic assumption that similar actions in different videos will have similar dynamics ($\mathcal{D}$). Namely, we assume there is a theoretical probability density function $p_{\mathcal{D}}$ based on which different instances of video-wide temporal evolutions are sampled for an action type. Naturally, different videos of the same action will be different and generate different ranking functions, so each linear ranker will have a different parametric representation vector $\mathbf{\psi}$. Therefore, a rightful question is to what extent learning the $\psi$ per video generalizes well for different videos of the same action.

As we cannot know the theoretical probability density function $p_{\mathcal{D}}$ of dynamics in real world videos, it is not possible to derive a strict bound on the generalization capacity of the functional parameters $u_i$. However, the sensitivity risk minimization framework gives us a hint of this generalization capacity of $\mathbf{u_i}$ when the input for the training is slightly perturbed. More specifically, Bousquet \etal.~\cite{BousquetJMLR2002} showed on a wide range of learning problems, \eg SVM, SVR and RankSVM, that the difference of the generalization risk $R$ from the leave one out error $R_i$ in the training set is bounded by
\begin{equation}  
|R - R^{/i}| \le E_r[|l(A_S,r)-l(A_{S^{/i}},r)|] \le \beta,
\label{eq:bound}
\end{equation}
where $A_S$ is a ranking algorithm with uniform stability $\beta$ learned from the set of samples $S$. The expectation of the loss over the distribution $r$ is denoted by $E_r[l]$ where $l$ is a bounded loss function such that $0 \le l(A_S,r) \le M$; ($M$ is a sufficiently small number).

Given a certain video, eq.~\eqref{eq:bound} implies that a slight change (ignoring smoothing of sequences) during training will learn a ranking function $\psi_{/i}$ with an error no larger than $\beta$ compared to the $\psi$ learned when all frames are available. Although eq.~\eqref{eq:bound} does not give a strict answer for what happens when the training input changes significantly from video to video, it hints that since the temporal evolution of similar actions should be similar, this should also be the case for the learned ranking functions of \name denoted by $\mathbf{u}$. This generalization capacity of \name is furthermore supported by our experimental validation.

\subsection{Functional parameters as temporal pooling}
\label{sec:parameter-pooling}

In the above we described how to encode temporal information from a video sequence using ranking machines.
The parameters $\mathbf{u}$ that we learn either from a pair-wise ranking machine or a point-wise ranking machine can be viewed as a principled, data-driven, temporal pooling method, as they summarize the data distributions over a whole sequence.
The use of ranking functional parameters as temporal pooling contrasts with other standard methods of pooling, such as \emph{max} pooling or \emph{sum} pooling, which are typically used either in convolutional neural networks~\cite{Krizhevsky2012} or for aggregating Fisher vectors~\cite{Perronnin2010}.

\begin{figure}[t!]
   \includegraphics[width=\linewidth]{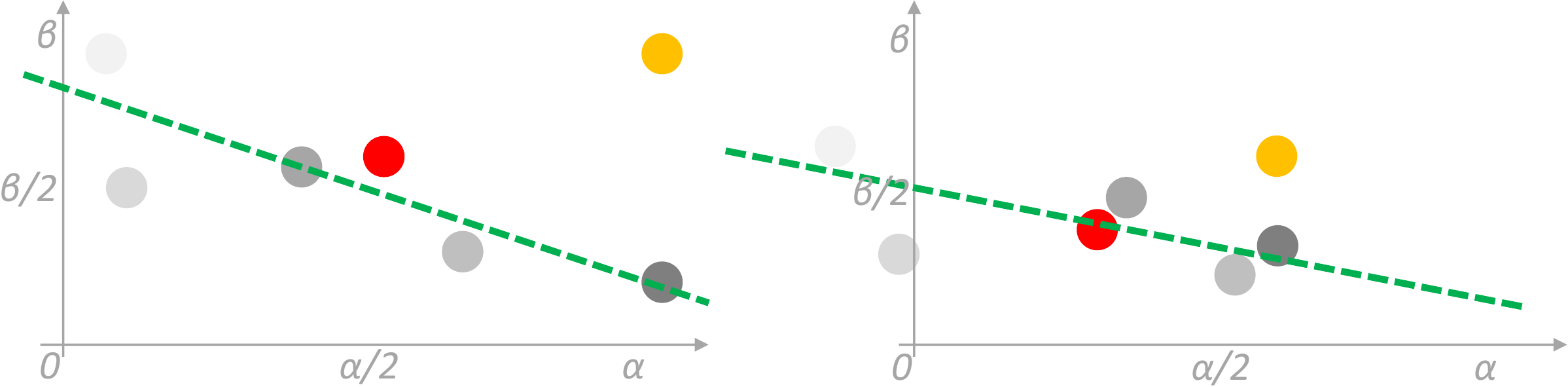}
    \caption{Various pooling operations given data plotted on a 2d feature space (gray circles stands for data, red circles for average pooling and yellow circles for max pooling, whereas the green dashed lines stand for rank pooling). The green dashed hyperplanes returned by our rank pooling not only describe nicely the latent data structure, but also are little affected by the random data noise. In contrast the average and max pooling represenations are notably disturbed. In fact, max pooling even creates ``ghost'' circles in areas of the feature space where no data exist.}
    \label{fig:pooling}
\end{figure}

First, as rank pooling is regularized, it is much less susceptible to the local noise in the observations \ie robust. See for example the left picture in Fig.~\ref{fig:pooling}, where max pooling is notably affected by stochastic perturbations in the feature space of the sequence data. Second, given some latent structure, temporal structure in our case, rank pooling fits the data trend by minimizing the respective loss function. Max pooling and sum pooling, on the other hand, are operators that do not relate to the underlying temporal data distribution. As such, max and sum pooling might aggregate the data by creating artificial, \emph{ghost} samples, as shown in the right picture of Fig.~\ref{fig:pooling}. In contrast, rank pooling transits the problem to a dual parameter space, in which the aggregation point is the one that optimally represents the latent data structure, as best expressed by the respective parametric model.

Next, we extend further the idea of using functional parameters as representations with different parameteric models.
Assume a function which learns a projection of the video frames into a subspace. Also, assume that we have enriched the frame representations so that they are more correlated with the time arrow, as we will discuss in Section~\ref{sec:action-videorepr}.
Then, another way to capture the video temporal evolution of appearances and the dynamics of the video would be to fit a function that reconstructs the time-sensitive appearance of all frames. 

To reconstruct the time-sensitive appearance of all frames in a video sequence $V$, we need to fit a function $\Psi(t, \mathbf{v_{1:t}; \mathbf{u}})$, such that
\begin{equation}
u^* = \argmin{u} \| V - u u^T V \|^2,
\label{eq:reconstruction}
\end{equation}
where $\mathbf{u} \in \Real^{D \times d}$, where $d$ is the new subspace dimensionality. In equation~\eqref{eq:reconstruction} we minimize the reconstruction error after a linear projection. One can solve the above minimization using principal component analysis, namely by singular value decomposition
\begin{equation}
V = U \Sigma U'^T
\label{eq:reconstruction2}
\end{equation}
The singular value decomposition returns two orthonormal matrices $U \in \Real^{D\times D}, U'\in \Real^{T\times T}$, who contain the eigenvectors of the covariance matrices $\hat{C}=E(V V^T)$ and $\hat{C'}=E(V^T V)$ respectively, where $T$ is the number of frames in the video sequence $V \in \Real^{ D \times T }$.

The straightforward way of defining the subspace $u$ is by selecting the $k$ first eigenvectors from $U$. However, more often than not the number of frames is smaller than the dimensionality of the frame features, $T < D$. Hence, the matrix $\hat{C}$, which is the expected value of the real but unknown covariance matrix $C$, is an unreliable estimate. To obtain a more robust subspace projection, we can instead consider
%
\begin{equation}
u= U' (V^T)^{-1}
\label{eq:pca2}
\end{equation}

Since $U'$ is obtained from the more robust estimate $\hat{C'}$, the subspace projection $u$ from eq.~\ref{eq:pca2} is a more reliable representation of the temporal evolution of the appearances in $V$. We can therefore use $u$ from eq.~\ref{eq:pca2} to represent the video $V$. Naturally, one can maintain only the first $d$ principal components of $U'$ to control the final dimensionality of $u$.

Given that frame features should ideally be correlated with the time variable, the first principal eigenvector contains the highest variance of the video appearance as it evolves with time. Therefore, one can use the first principal component of a video as a temporal representation given that video frames are pre-processed to indirectly correlate with time (see Section~\ref{sec:action-videorepr}). We refer to the above functional parameter pooling as \textbf{\emph{subspace pooling}}. Subspace pooling is robust, as also shown in~\cite{Fernando2013b}. Moreover, the subspace pooling has a close relationship to dynamic texture\cite{Doretto2003} which uses auto-regressive moving average process which estimates the parameters of the model using sequence data. Subspace pooling is also related to dynamic subspace angles\cite{Li2011} which compares videos by computing subspaces and then measuring the principal angle between them.

Support vector ranking machines and principal component decomposition are robust models which we can use for temporal pooling.
However, other learning algorithms can also be considered to be used as functional parameter representation.
As a case study, in this paper we use two more popular choices: \emph{(a)} Hidden Markov Models (HMMs), and \emph{(b)} Neural Networks (NN). For details, we refer to the experimental results section. We explore the different possibilities experimentally.


\section{Frame representations}
\label{sec:action-videorepr}
Even in a noise-free world video data would still exhibit high degrees of variability. To reduce the effect of noise and violent abrupt variations, we smooth the original video signal (\ie the frame representation $\mathbf{x_t}$). In this section we discuss three methods to obtain smoothed robust signals $\mathbf{v_t}$ from frame data $\mathbf{x_t}$.    

\begin{figure}[t!]
    \centering {
    
\subfloat[]{\includegraphics[width=0.33\linewidth]{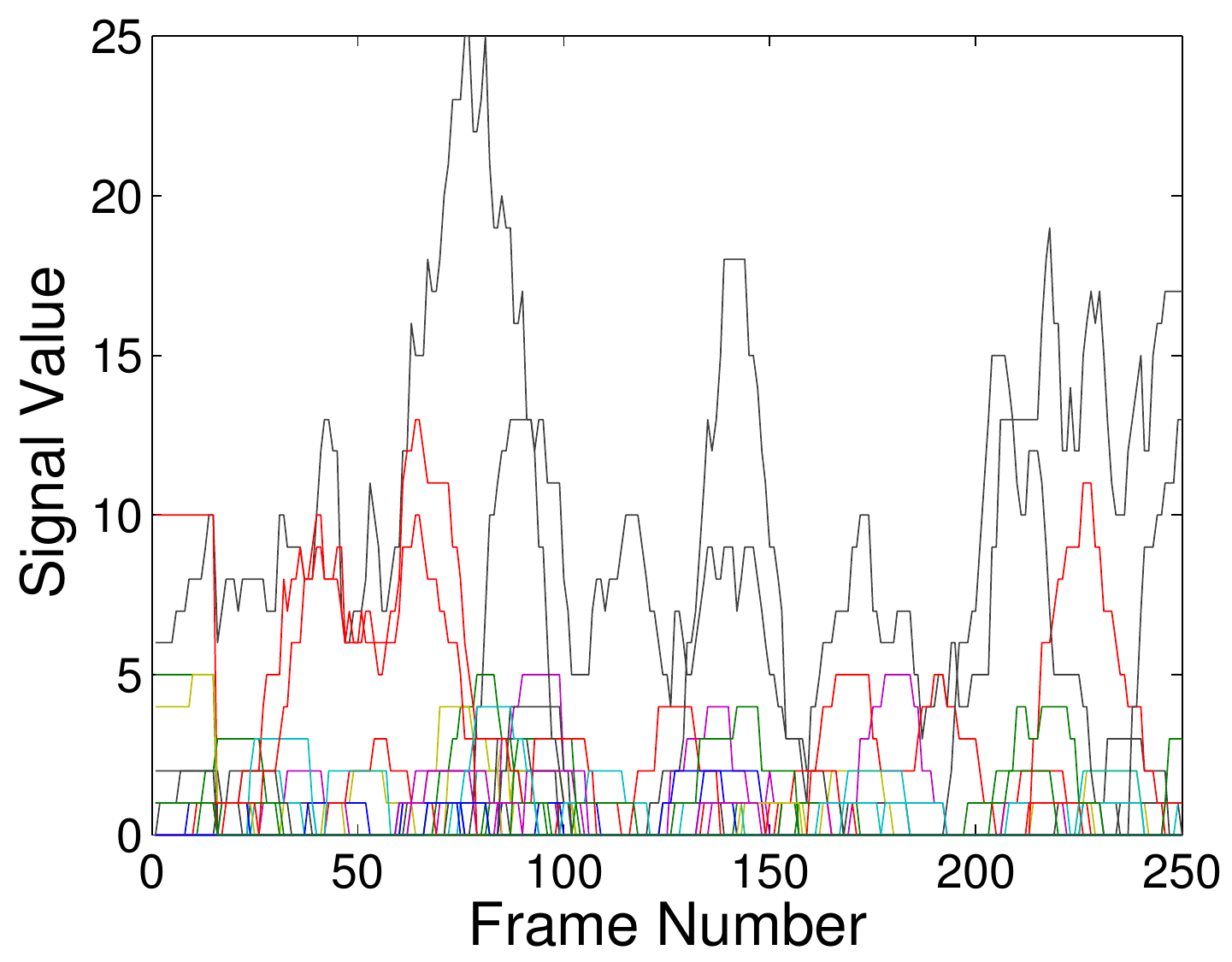}}
    \subfloat[]{\includegraphics[width=0.33\linewidth]{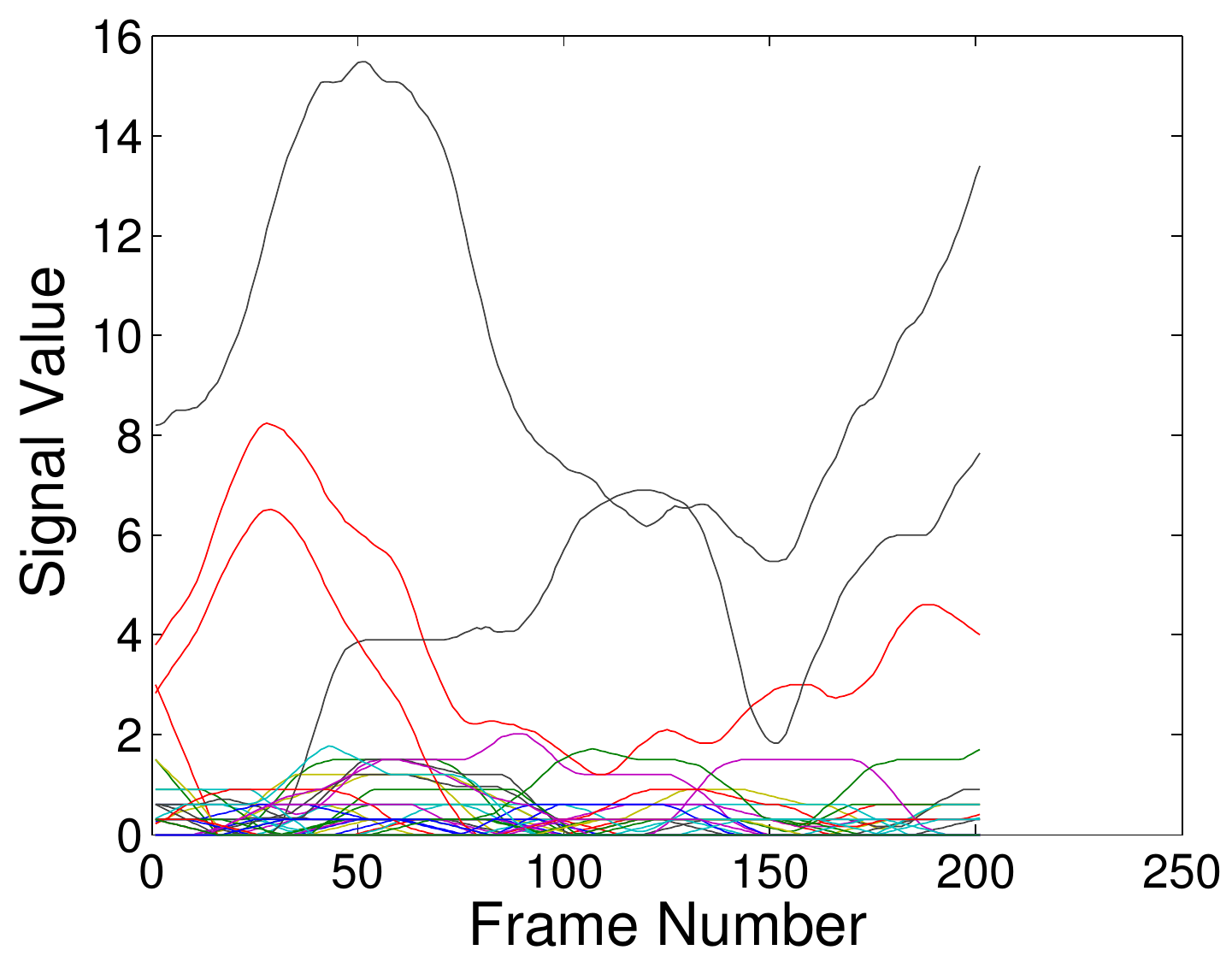}}
    \subfloat[]{\includegraphics[width=0.33\linewidth]{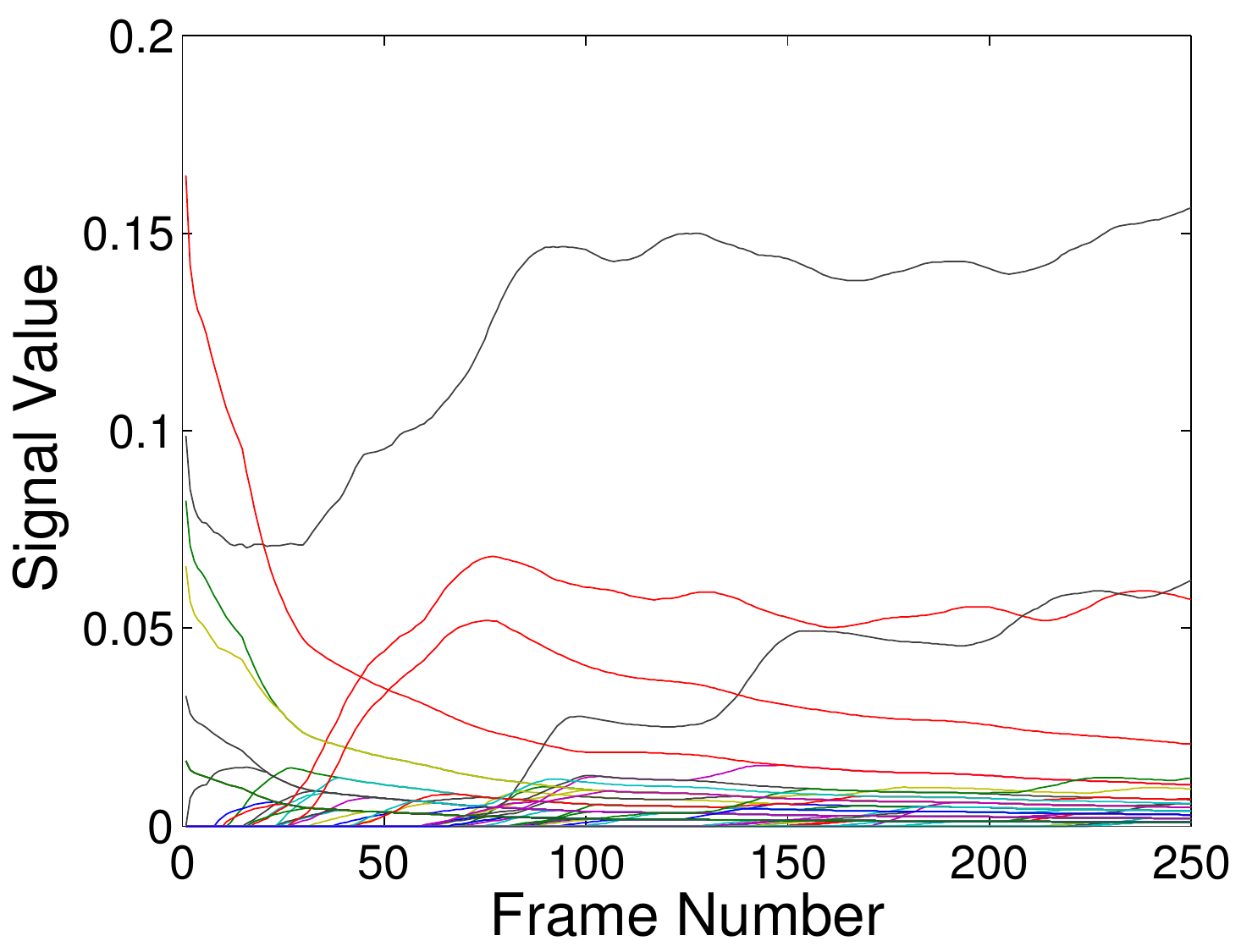}} \\
    
\subfloat[]{\includegraphics[width=0.33\linewidth]{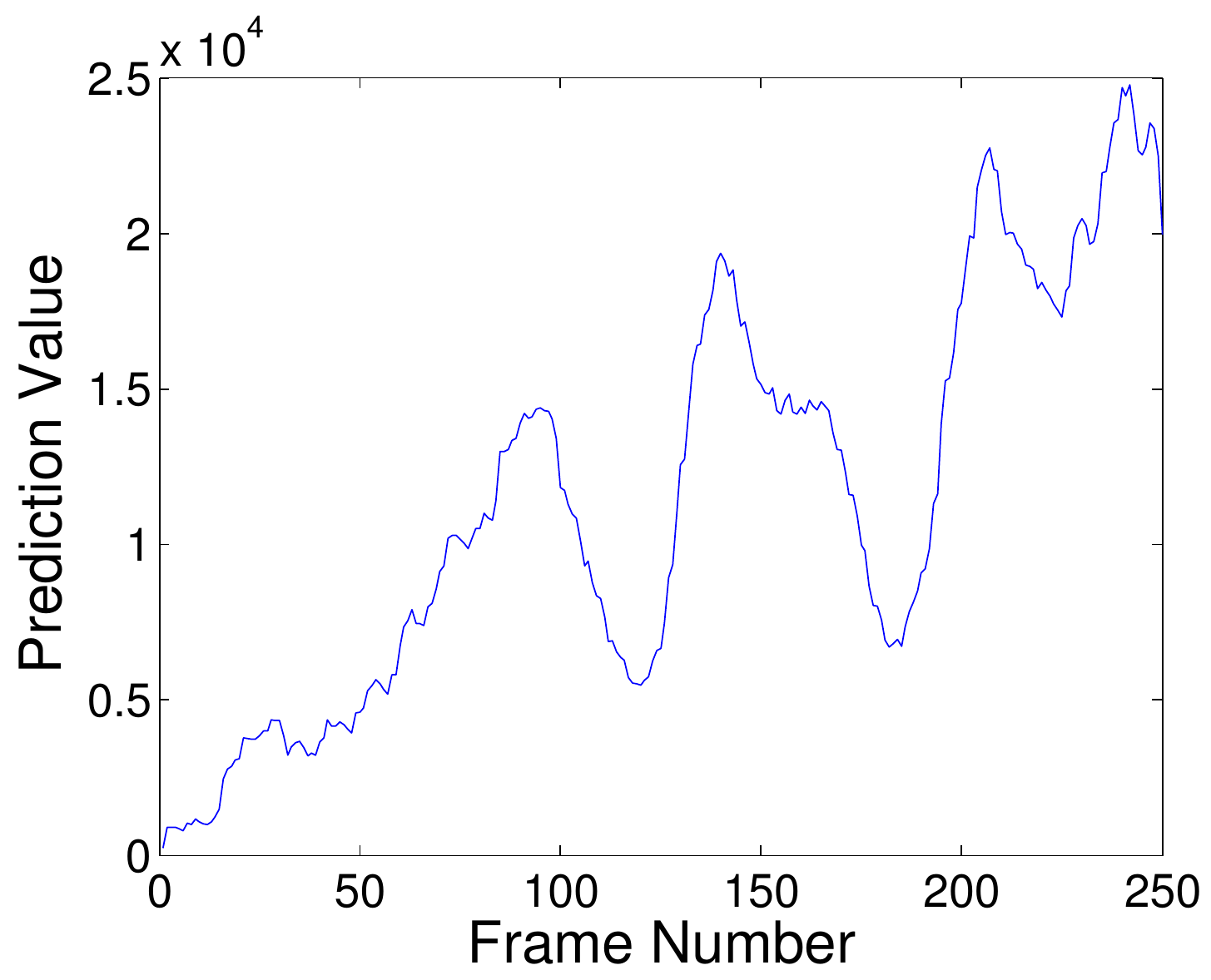}}    
\subfloat[]{\includegraphics[width=0.33\linewidth]{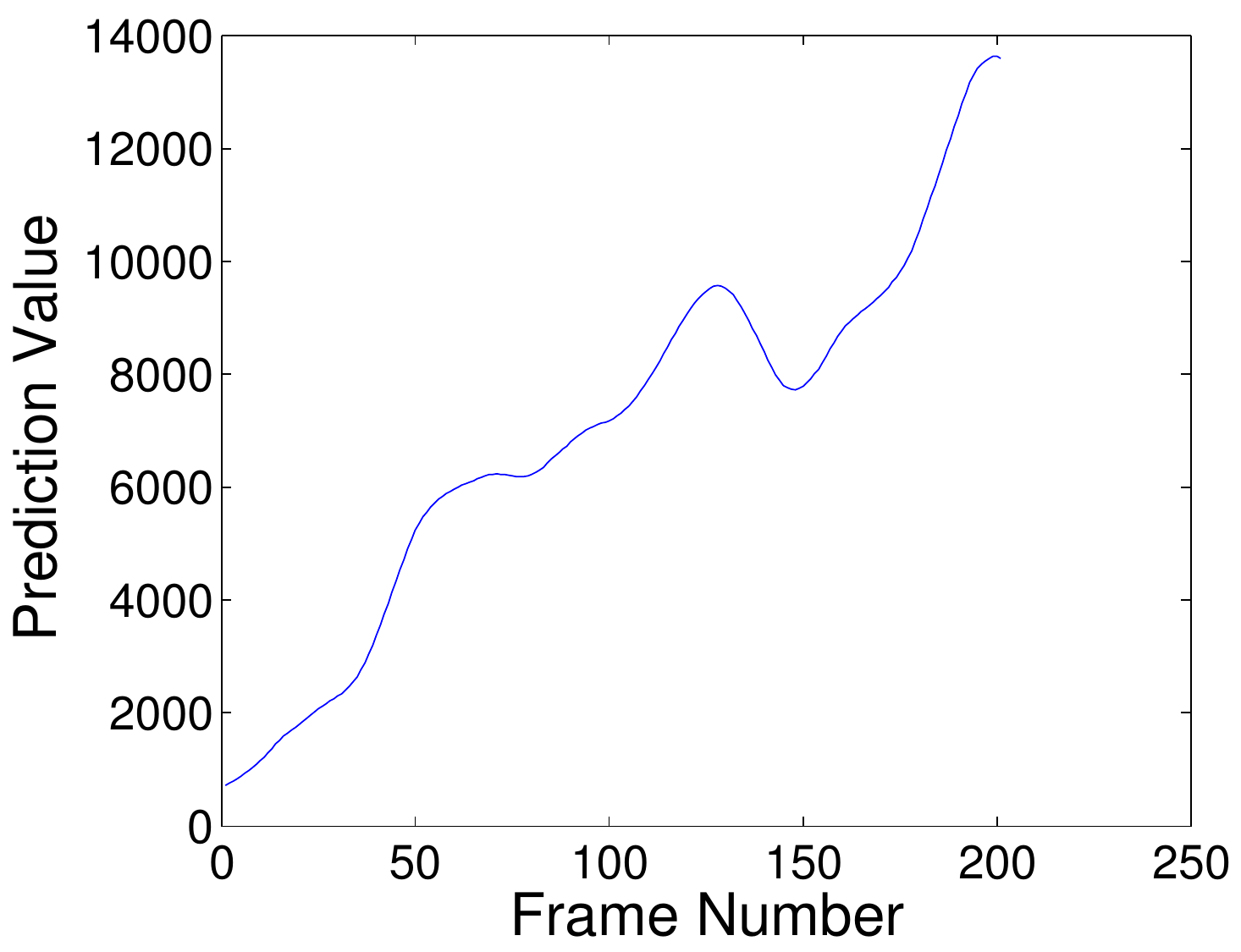}}    
\subfloat[]{\includegraphics[width=0.33\linewidth]{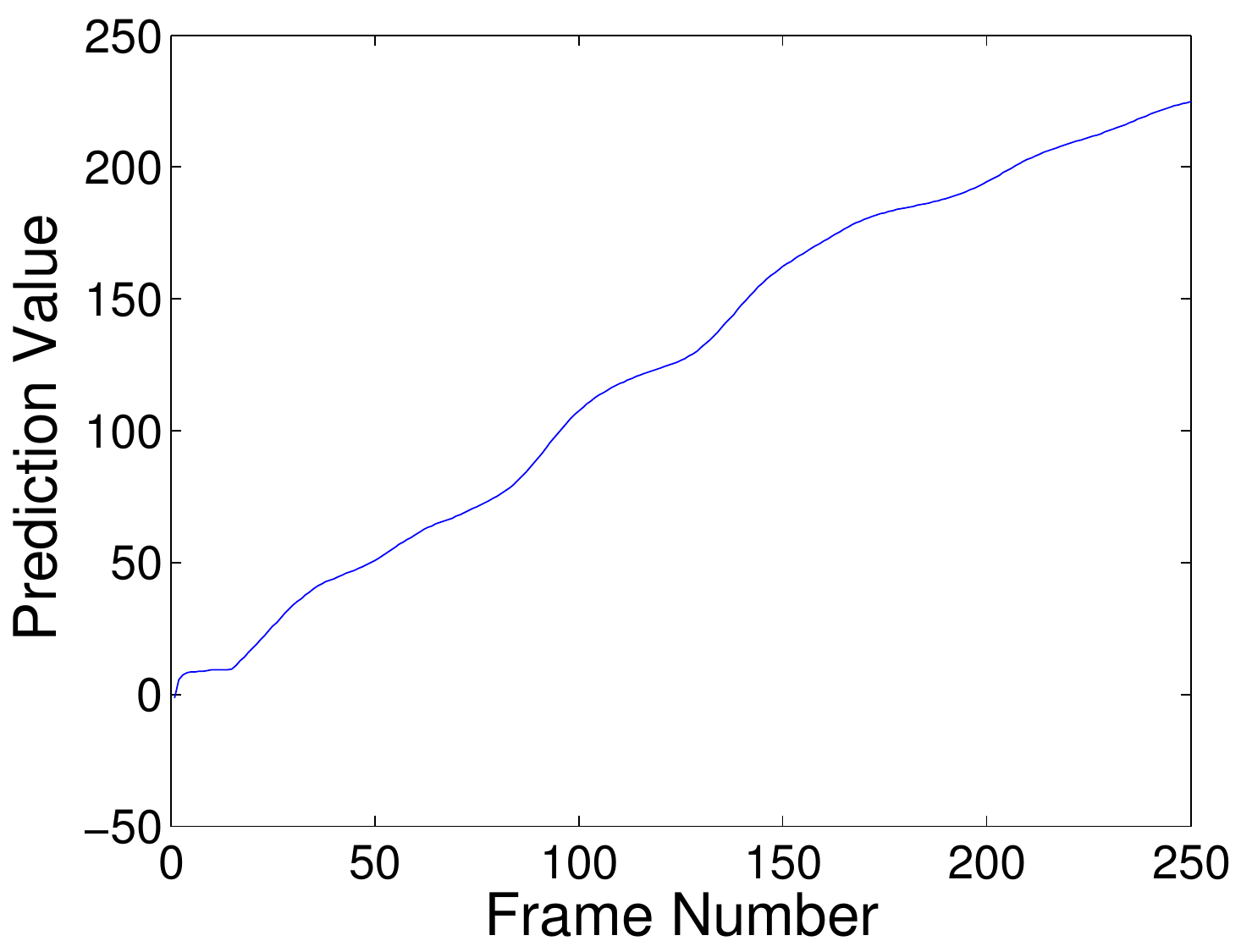}}
    }
     \caption{
Using ranking machines for modeling the video temporal evolution of appearances, or alternatively, the video dynamics. We see in (a) the original signal of independent frame representation, (b) the signal obtained by moving average, (c) the signal obtained by time varying mean vector (different colors refer to different dimensions in the signal $\mathbf{v_t}$). In (d), (e) and (f) we plot the predicted ranking score of each frame obtained from signal (a), (b) and (c) respectively after applying the ranking function (predicted ranking value at t, $\mathbf{s_t}= \mathbf{u}^T  \cdot \mathbf{v_t}$).}
    \label{fig:representation-dynamics}
\end{figure}

\subsection{Independent Frame Representation}

The most straightforward representation for capturing the evolution of the appearance of a video is to use independent frames  $\mathbf{v_t}= \frac{\mathbf{x_t}}{\|\mathbf{x_t}\|}$. This approach has two disadvantages. First, the original signal can vary significantly, see Figure~\ref{fig:representation-dynamics}(a), often leading  the ranking machines to focus on undesirable temporal patterns. At the same time independent frames might generate ranking functions with high ranking errors during training time. Second, independent frame representations are characterized by a weak connection between $\mathbf{v_t}$ and $t$. Given this weak correlation between the $\mathbf{v_t}$ and time $t$, see Figure.~\ref{fig:representation-dynamics} (a), the ranking function may not learn the appearance evolution over time properly. As a result, plotting the predicted score $\mathbf{s_t}=\mathbf{u_i}^T \cdot \mathbf{v_t}$ for each of the frames in the video is not as smooth as one would desire (see Figure~\ref{fig:representation-dynamics} (d)).

\subsection{Moving Average~(MA)}

Inspired by the time series analysis literature, we consider the moving average with a window  size $T$ as video representation at time $t$. In other words we consider locally smoothed signals.
For MA, we observe two facts. First, the output signal is much smoother, see Figure~\ref{fig:representation-dynamics}(b). Second, $\mathbf{v_t}$ maintains a temporally local dependency on the surrounding frames around $t$, namely the frames $[t,t+T]$. Unlike the independent frames representation, however, the moving average model forges a connection between $\mathbf{v_t}$ and $t$. Plotting these two variables for a window T=50 in Figure~\ref{fig:representation-dynamics}(b), we observe a smoother relation between the dimensions of $\mathbf{v_t}$ and the frame number which equals to the time variable. As such, the video-wide temporal information is captured well in the predicted score $s_t$, see Figure~\ref{fig:representation-dynamics}(e).

Although the moving average representation allows for capturing the appearance evolution of a video better, we still witness a general instability in the signals. Furthermore, we note that the moving average representation introduces undesirable artifacts. For one, window size $T$ has to be chosen, which is not always straightforward as actions often take place in different tempos. Moreover, due to boundary effects, $v_t$ is undefined for the last time stamps $t$ of the video.

\subsection{Time Varying Mean Vectors}

To deal with the limitations of the independent frames representation and the moving average, we propose a third option, the \emph{time varying mean vectors}.

\begin{figure*}[t!]
    \centering 
    \includegraphics[width=1.0\linewidth]{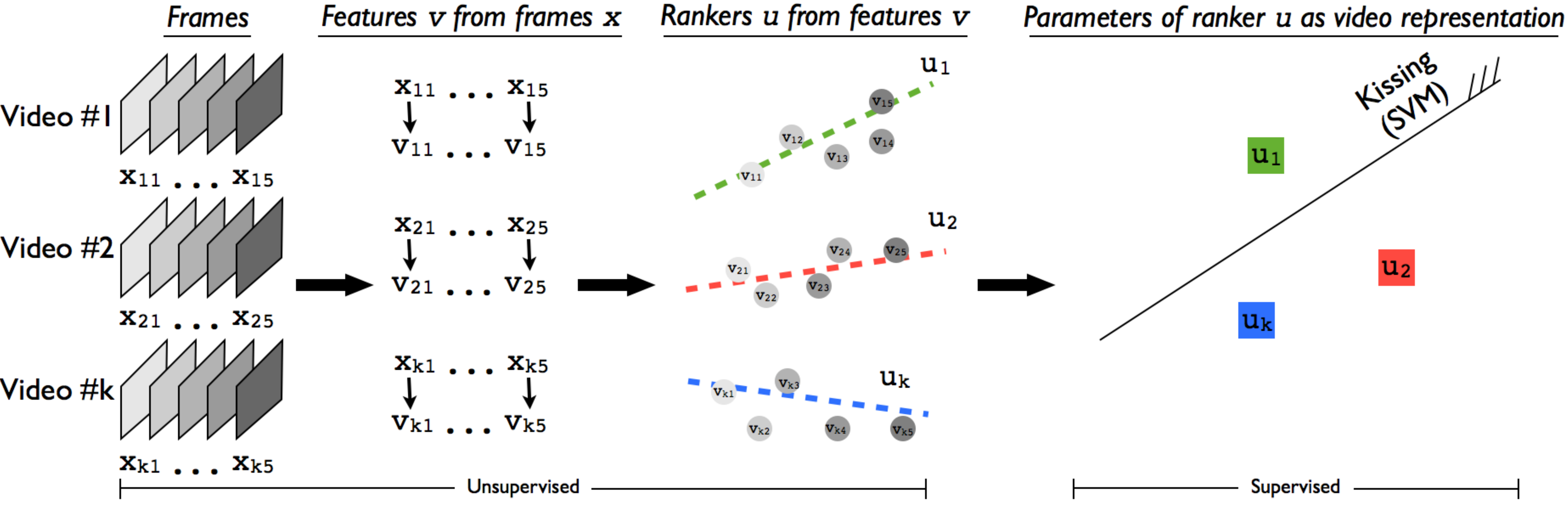}   
     \caption{Processing steps of \name for action recognition. First, we 
extract frames $\mathbf{x_1} \ldots \mathbf{x_n}$ from each video. Then we 
generate feature $\mathbf{v_t}$ for frame $t$ by processing frames from 
$\mathbf{x_1}$ to $\mathbf{x_t}$ as explained in 
section~\ref{sec:action-videorepr}. Afterwards, using ranking machines we learn 
the video representation $\mathbf{u}$ for each video. Finally, video 
specific $\mathbf{u}$ vectors are used as a representation for 
action classification.}
    \label{fig:mainaction}
\end{figure*}

Let us denote the mean at time $t$ as $\mathbf{m_t} =\frac{1}{t} \times \sum_{\tau=1}^{t} \mathbf{x_\tau} $. Then, $\mathbf{v_t}$ captures only the direction of the unit mean appearance vector at time $t$, \ie ($\mathbf{v_t} = \frac{\mathbf{m_t}}{||\mathbf{m_t}||}$). Thus the ranking function $\psi$  learns the evolution of the normalized mean appearance at time $t$. We plot the relationship between $\mathbf{v_t}$ and $t$ in Figure~\ref{fig:representation-dynamics}(c) and the prediction score $s_t$ in Figure~\ref{fig:representation-dynamics}(f). We observe that, as desired, the output is smooth, almost resembling a monotonically increasing function. Different from the independent frames representation, the time varying mean vectors introduce a better dependency between the input $v_t$ and the target $t$.

By construction \textit{time varying mean vectors} capture only the temporal information from the forward flow of the video with respect to the time. This is because the video progresses from the past to the future frames. However, there is no reason why the mean vectors should not be considered also in the reverse order, starting from the future frames and traversing backwards to the past frames of a video. To this end we generate the exact same objective, as in eq.~\ref{eq:dynamicsrank}, playing the video in reverse order, however. We shall refer to appearance evolution captured by forward flow as \textit{forward \name} (\textbf{FDRP}), whereas reverse flow as \textit{reverse \name}.\\

\subsection{Non-linear \name}
\label{sec.non.lin.rp}
In section~\ref{sec:rankpooling}, we considered only linear machines to obtain rank pooling based video representations.
To incorporate non-linearities we resort to non-linear feature maps~\cite{Vedaldi2012} applied on each $\mathbf{v_t}$ of $V$, thus allowing for employing effective~\cite{GavvesCVPR12} linear ranking machines in their primal form.

A popular technique to include non-linearities is to pre-process and transform the input data by non-linear operations. Let us denote a point-wise non-linear operator $\Phi(\cdot)$ which operates on the input $x$ so that the output $\Phi(x)$ is a non-linear mapping of $x$. We use such non-linear feature maps to model non-linear dynamics of input video data. Given the time varying mean vector $\mathbf{v_t}$, to obtain non-linear representation $\mathbf{u}$ of input video $X$, we map $\mathbf{v_t}$ to $\Phi(\mathbf{v_t})$ using the non-linear operation before learning the ranking machines. Next we describe an interesting non-linear feature map that is useful particularly for real data such as Fisher vectors. In our experiments we also demonstrate the advantage of capturing non-linear dynamics via non-linear feature maps which we coined non-linear \name.

A popular kernel in visual recognition tasks is the Hellinger kernel
\begin{equation}
K_{hell}(x, y) = \sqrt{x}^T \sqrt{y}.
\label{eq:hellinger}
\end{equation}
The Hellinger kernel introduces non-linearities to the kernel machines, while maintaining separability, thus allowing for solving the optimizations in their primal form.
The Hellinger kernel copes well with the frequently observed feature burstiness~\cite{Arandjelovic2012}. When eq.~\eqref{eq:hellinger} is applied directly, then we obtain a complex kernel, as the negative features turn into complex numbers, namely we have that $\sqrt{x} = \sqrt{x^+} + i \sqrt{x^-}=\hat{x}^+ + i \hat{x}^-$, where $\hat{x}^+=\sqrt{x^+}$ and $\hat{x}^-=\sqrt{x^-}$ refer to the positive and negative parts of the feature $x$, namely $x^+_i=x_i, \forall x_i>0$ and 0 otherwise, while $x^-_i=-x_i, \forall x_i<0$ and 0 otherwise. Then the Hellinger kernel equals to
\begin{eqnarray}
K_{hell}(x, y) & = & (\hat{x}^+ + i \hat{x}^-)^T (\hat{y}^+ + i \hat{y}^-)  \\ \nonumber
               & = & (\hat{x}^+ \hat{y}^+ - \hat{x}^- \hat{y}^-) + i (\hat{x}^- \hat{y}^+ + \hat{y}^- \hat{x}^+)
\label{eq:complex}
\end{eqnarray}

To avoid any complications with using complex numbers, we focus on the real part of $K_{hell}$.
Using the real part of the Hellinger kernel, we effectively separate the positive and negative parts of the features, easily deriving that
\begin{eqnarray}
K_{\operatorname{Re}\{hell\}} & = & [\hat{x}^+, \hat{x}^-] [\hat{y}^+, \hat{y}^-]^T \nonumber \\ 
                                    & = & K_{hell}(x^*, y^*), \label{eq:complex2}
\end{eqnarray}
where $x^*=[x^+, x^-]^T$ is the \emph{expanded} feature, which is double in dimensionality compared to $x$ and is composed of only positive elements. Comparing eq.~\eqref{eq:complex2} with eq.~\eqref{eq:hellinger}, we observe that we have practically doubled the dimensionality of our feature space, as all $x, \hat{x}^+, \hat{x}^-$ have the same dimensionality, allowing for more sophisticated learning. The first half of the feature space relates to the positive values only $\hat{x}^+$, while the second part relates to the negative ones $\hat{x}^-$. We refer to this feature map as \emph{posneg} feature map and, to the respective kernel as \emph{posneg} kernel. Unless stated otherwise, in the remainder of the text we use the posneg feature maps.


\section{Overview}
\label{sec:action-overview}

\begin{figure*}[t!]
    \centering
    \includegraphics[width=0.9\linewidth]{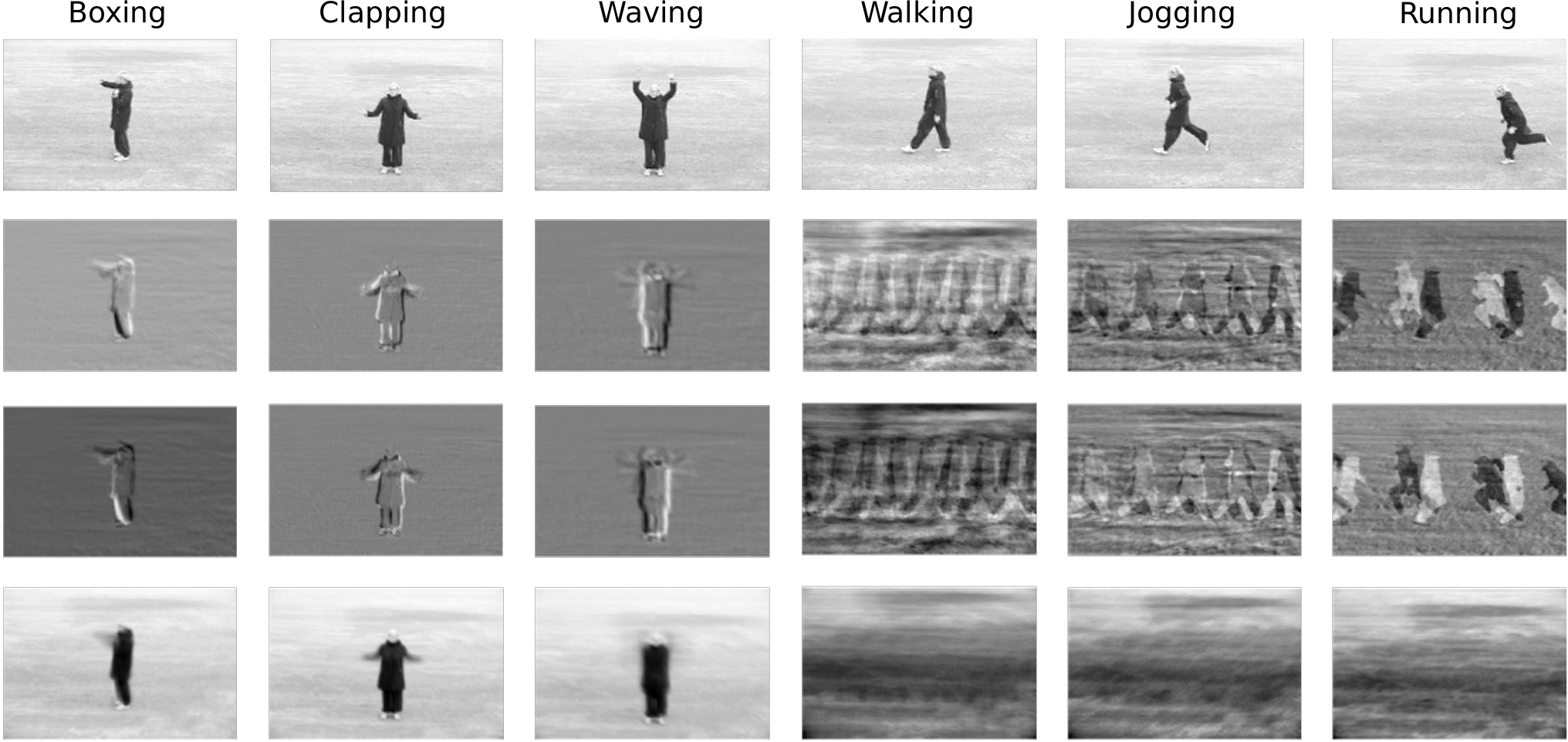}
    \caption{Examples from the six action categories in the KTH action recognition dataset~\cite{LaptevL03}. From left to right you see the actions \emph{boxing}, \emph{clapping}, \emph{waving}, \emph{walking}, \emph{jogging} and \emph{running}. From top to bottom you see an example frame from a random video, the forward \name, the reverse \name and the result after the standard mean pooling. The \name as well as the mean representations are computed on the image pixels. We observe that the forward and reverse \name indeed capture some of the crisp, temporal changes of the actions, whereas the mean representations lose the details.}
    \label{fig:action-KTH}
\end{figure*}

Next, we will briefly describe the pipeline for applying \name for the task of action classification in videos.

\subsection{Action classification from A to Z} 

The action classification pipeline is illustrated in detail in Figure~\ref{fig:mainaction}. First, for each video $X_i$ the video frames $x_{ij}, j=1, \dots M$ are processed individually, so that frame feature encodings $v_{ij}$ are extracted and their frame location in the video is recorded. A popular choice to date would be to first extract HOG, HOF, MBH, TRJ features computed on improved trajectories per frame together with the frame location, then compute the per frame Fisher vector or the Bag-of-Words feature encodings (first two columns in Figure~\ref{fig:mainaction}). Next, given the smooth frame features obtained from time varying mean vectors or any of the other frame representations discussed in Section~\ref{sec:action-videorepr}, we apply a parametric pooling step. For each of the videos $X_i$ we fit one of the parametric models discussed in Section~\ref{sec:action-theory} (third column in Figure~\ref{fig:mainaction}). We then use the parameters $u_i$ of the parametric model as the video representation. Last, after having computed all $u_i$ for every video $X_i$, we run a standard supervised classification method on our dataset denoted by $D_{train}=\{\mathbf{u_i}, y_i\}, i=1, \dots, N$ where $N$ is the number of videos in our training set and $y_i$ is the class label of the $i^{th}$ video. We use non-linear SVM classifiers such as $\chi^2$ feature maps~\cite{vedaldi08vlfeat} applied on feature vectors $\mathbf{u_i} \in \Real^D$.

We summarize some of the advantages of using parameters of a function that is trained to map or correlate input data to time variable as a video representation. First, no supervised information is needed as video order constraints can be obtained directly from the sequence of video frames. Second, by minimizing eq.~\eqref{eq:dynamics} \name captures the evolution of appearance of a video in a principled manner, either by minimizing a ranking objective or by minimizing the reconstruction error of video appearances over time. Third, such a parametric representation does not require negative data to be added explicitly during the learning of the video representations. Fourth, since \name encapsulates the changes that occur in a video, it captures useful information for action recognition.


%
\subsection{Visualising dynamics of videos}
\label{sec.motion.images}

In this section we demonstrate a visual inspection of what our \name method learns. For simplicity of visualization we use sample video sequences from the KTH action recognition dataset~\cite{LaptevL03}. As features we use the raw RGB values vectorized per frame as features. In this visualization experiment we do not extract any trajectories or other more sophisticated features and we use independent frame representations. We apply forward and reverse \name on the video sequences of the first row. To obtain the visualization, given a frame image we first transform it to a D-dimensional gray-scaled vector. Then we apply the \name method to obtain the parameters $\mathbf{u}$. Afterwards, we reshape the vector $\mathbf{u}$ to the original frame image size and project back each pixel value to be in the range of 0-255 using linear interpolation by min-max normalization.

We use example videos provided in the dataset which consists of six action classes, namely boxing, hand clapping, hand waving, walking, jogging and running. Samples from this dataset are shown in the first row of Figure~\ref{fig:action-KTH}. For each of the 6 actions in the KTH dataset we present a sample sequence in each column of Figure~\ref{fig:action-KTH} (from left to right we have \emph{boxing}, \emph{clapping}, \emph{waving}, \emph{walking}, \emph{jogging} and \emph{running}). In the second and third row we show the forward and reverse \name video representation respectively (namely the computed $u_i$), illustrating the captured temporal motion information. In the last row we show the result of the standard average pooling. When the motion of the action is apparent, \name method seems to capture this well. What is more interesting is that, not only \name separates running in one direction from the other direction, but also seems to capture the periodicity of the motion to an extent, see the last column of Figure~\ref{fig:action-KTH} that depicts \emph{running}. 


\section{Experiments}
\label{sec:action-exp}
Now we present a detailed experimental evaluation of \name.\\


\begin{figure}[t]
\centering
 \includegraphics[width=0.98\linewidth]{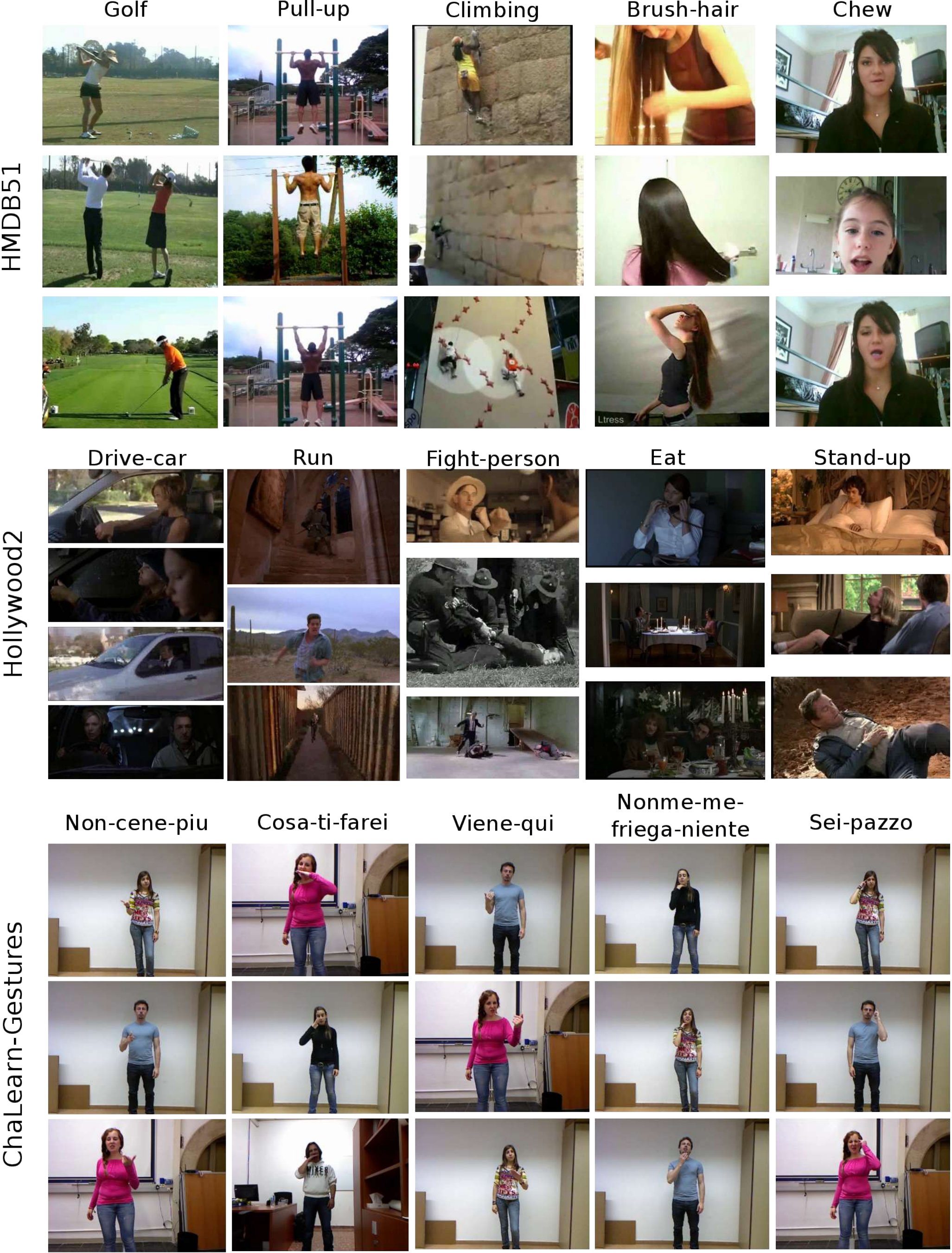}
 \caption{Some example frames from the top performing categories of the HMDB51, Hollywood2, and ChaLearn-Gestures dataset, respectively.}
 \label{fig.hmdb.examples}
\end{figure}

\noindent\textbf{Datasets.} As the proposed methodology is not specific to an action type or class of actions, we present experiments in a broad range of datasets. 
We follow exactly the same experimental settings per dataset, using the same training and test splits and the same features as reported by the state-of-the-art methods.\\

\noindent\textbf{HMDB51 dataset~\cite{Kuehne2011}}. This is a generic action classification dataset composed of roughly 7,000 clips divided into 51 action classes. 
Videos and actions of this dataset are subject to different camera motions, viewpoints, video quality and occlusions. 
As done in the literature we use a one-vs-all classification strategy and report the mean classification accuracy over three standard splits provided by the authors in~\cite{Kuehne2011}.
Some example frames from this challenging dataset are shown in Figure~\ref{fig.hmdb.examples}.\\

\noindent\textbf{Hollywood2 dataset~\cite{Laptev2008}} This dataset has been collected from 69 different Hollywood movies that include 12 action classes. 
It contains 1,707 videos in total where 823 videos are used for training and 884 are used for testing. 
Training and test videos are selected from different movies. 
The performance is measured by mean average precision (mAP) over all classes, as in \cite{Laptev2008}. \\

\noindent\textbf{MPII cooking activities dataset~\cite{rohrbach2012database}}. 
This dataset was created to evaluate fine-grained action classification. 
It is composed of 65 different actions that take place continuously within 8 hours of recordings. 
As the kitchen remains the same throughout the recordings, the classification focuses mainly on the content of the actions and cannot benefit from potentially discriminative background information (\eg driving a car always takes place inside a car). 
We compute per class average precision using the same procedure as in ~\cite{rohrbach2012database} and report the final mAP. \\

\noindent\textbf{ChaLearn Gesture Recognition dataset~\cite{Escalera2013}}. 
This dataset contains 23 hours of Kinect data of 27 persons performing 20 Italian gestures. 
The data includes RGB, depth, foreground segmentation and Kinect skeletons. 
The data is split into train, validation and test sets, with in total 955 videos each lasting 1 to 2 minutes and containing 8 to 20 non-continuous gestures. 
As done in the literature, we report precision, recall and F1-score measures on the validation set. \\

\noindent\textbf{\Name and baselines.} In Sec.~\ref{sec.action.exp.rep} and ~\ref{sec.action.exp.stability} we compare different variants of \name. 
As a first baseline we use the state-of-the-art trajectory features (\ie improved trajectories and dense trajectories) and pipelines as in~\cite{wang2013action, wang2013dense}. 
As this trajectory-based baseline mainly considers \textit{local temporal information} we refer to this baseline as \textit{local}. 
We also compare with temporal pyramids (\textit{TP}), by first splitting the video into two equal size sub-videos, then computing a representation for each of them like spatial pyramids~\cite{Lazebnik2006}. 
For these baselines, at frame level we apply non-linear feature maps (\ie power normalization for Fisher vectors and chi-squared kernel maps for bag-of-words-based methods). 
We also compare different versions of \name, we denote the forward \name by \textit{FDRP}, the reverse \& forward \name by \textit{RFDRP}, the non-linear forward \name by \textit{NL-FDRP} and the non-linear reverse \& forward \name by \textit{NL-RFDRP}.\\

\noindent\textbf{Implementation details.}  In principle there is no constraint on the type of linear ranking machines we employ for learning \name. 
We have experimented with state-of-the-art ranking implementation RankSVM~\cite{JoachimsKDD2006} and SVR~\cite{Smola2004}. 
Both these methods can be used to solve learning to rank problems formulated in equation~\ref{eq:dynamicsrank}. 
We observe that both methods capture evolution of the video appearances equally well. 
As for SVR the learning convergence is notably faster, we will use the SVR solver of Lib-linear in this paper (C = 1).

For HMDB51 and Hollywood2 datasets we use state-of-the art improved trajectory features~\cite{wang2013action} with Fisher encoding \cite{Perronnin2010}. 
As done in the literature, we extract HOG, HOF, MBH, and trajectory (TRJ) features from the videos. 
We create GMMs of size 256 after applying PCA with a dimensionality reduction of factor 0.5 on each descriptor. 
As done in ~\cite{wang2013action}, we also apply the square-root trick on all descriptors except for TRJ.

In order to compute non-linear \name, we apply features maps \emph{(posneg)} followed by a L2-normalization on individual Fisher vectors extracted from each video frame. 
For linear \name, we just use Fisher vectors without any power normalization. 

For MPII cooking dataset we use the features provided by the authors~\cite{rohrbach2012database}, that is bag-of-words histograms of size 4000 extracted from dense trajectory features~\cite{wang2013dense} (HOG, HOF, MBH and TRJ). 
As we use bag-of-words for this dataset, in order to compute non-linear \name, we apply $\chi^2$-kernel maps on individual bag-of-words histograms after the construction of the vector valued function as explained in section~\ref{sec:action-videorepr}.

For the ChaLearn Gesture Recognition dataset we start from the body joints~\cite{ShottonCVPR11}. 
For each frame we calculate the relative location of each body joint w.r.t. the torso joint. 
Then, we scale these relative locations in the range [0,1].
We use a dictionary of 100 words to quantize these skeleton features.  
Similar to MPII cooking dataset, in order to compute non-linear \name and for all baselines we use chi-squared kernel maps.

We train non-linear SVM classifiers with feature kernel maps for the final classification. 
Whenever we use bag-of-words representation we compute $\chi^2$-kernel maps over the final video representation and then L2 normalize them. 
We use this strategy for both baselines and \name.  
Similarly, when Fisher vectors are used, we use posneg feature map and L2 normalization for the final video representation. 
The C parameter of SVM is cross-validated over the training set using two-fold cross-validation to optimize the final evaluation criteria (mAP, classification accuracy or F-score). 
When features are fused (combined) we use the average kernel strategy.
We provide code for computing \name in a public website~\footnote{The code for computing \name, as well as scripts for running experiments for the different datasets can be found in http://bitbucket.org/bfernando/videodarwin.}.\\

\noindent\textbf{Execution time.} \Name takes about 
$0.9 \pm 0.1$ sec per video on the Hollywood2 dataset excluding the Fisher vector computation. The proposed algorithm is linear on the length of the video.

\subsection{\Name: Frame representations \& encodings}
\label{sec.action.exp.rep}

We first evaluate the three options presented in 
Section~\ref{sec:action-videorepr} for the frame representation, \ie 
\textit{independent frame}, \textit{moving average} and \textit{time varying 
mean vector} representations. We perform the experiments with Fisher vectors on 
the Hollywood2 dataset and summarize the results in 
Table~\ref{tab:video-representations}. Similar trends were observed with dense 
trajectory features, bag-of-words and other datasets.

From the comparisons, we make several observations that validate our analysis. 
First, applying ranking functions directly on the Fisher vectors from the frame 
data captures only a moderate amount of the temporal information. Second, 
\textit{moving average} applied with ranking seems to capture video-wide temporal 
information better than applying ranking functions directly on the frame data. 
However, the \textit{time varying mean vector} consistently outperforms the other 
two representations by a considerable margin and for all features. We believe 
this is due to two reasons. First, \textit{moving average} and \textit{time 
varying mean vector} methods smooth the original signal. This reduces the noise 
in the signal. Therefore, it allows the ranking function to learn meaningful VTE.
Secondly, the appearance information 
of the \textit{time varying mean vectors} is more correlated with the time variable.
The ranking function exploits this correlation to learn the evolution of the 
appearance over time in the video signal. 

We conclude that \textit{time varying mean vectors} are better for capturing the video-wide evolution of appearance of 
videos when applied with \name. In the rest of the experiments we use the time varying mean vectors.

Last, we evaluate the contribution of the time-varying mean vectors when used along with other pooling methods such as average pooling.
We perform an experiment on Hollywood2 using the MBH features. 
The average pooling on top of time-varying mean vectors gives an improvement of 0.5\% (relative to average pooling on FV directly) only, indicating that for average pooling, there is no advantage of time varying mean vectors.

\begin{table}[t]
  \centering
  \setlength{\tabcolsep}{6pt}
  \scalebox{0.80}{
  {\scriptsize
  \begin{tabular}{l c c c c c c}
  \toprule  
                    & HOG & HOF & MBH & TRJ & Comb. \\
  \midrule
  \textit{Independent frames}&
  41.6 & 52.1 & 54.4 & 43.0 & 57.4 \\
  \textit{Moving average (T=20)}&
  42.2 & 54.6 & 56.6 & 44.4 & 59.5\\
  \textit{Moving average (T=50)}&
  42.2 & 55.9 & 58.1 & 46.0 & 60.8 \\
  \textit{Time varying mean vectors} &
  \textbf{45.3} & \textbf{59.8} & \textbf{60.5} & \textbf{49.8} & \textbf{63.6}\\  
  \bottomrule
  \end{tabular}
  }
  }
  \caption{Comparison of different video representations for \name. Results 
reported in mAP on the Holywood2 dataset using FDRP with Fisher vectors. As also 
motivated in Sec.~\ref{sec:action-videorepr}, the time varying mean vector 
representation captures better the video-wide temporal information present in a 
video.}
  \label{tab:video-representations}
\end{table}

\subsection{Action classification}
\label{sec.action.exp.action}

Next, we present a detailed analysis of the action classification results in 
HMDB51, Hollywood2, MPII Cooking and ChaLearn Gesture recognition datasets 
(see Table~\ref{tab:HMDB51}, ~\ref{tab:holywood2}, ~\ref{tab:cooking} and 
~\ref{tab:gesture} respectively).

\Name obtains  better results in comparison to local temporal 
methods. Accurately modeling the evolution of appearance and motion, allows to 
capture more relevant information for a particular action. These results 
confirm our hypothesis that what makes an action most discriminative from other 
actions is mostly the video-wide evolution of appearance and motion information 
of that action. The forward and reverse \name variant reports consistent improvement 
over forward-only \name, further improved when non-linear \name is employed. It 
is interesting to see that this trend can be observed in all four datasets too. 
Overall, local methods combined with \name bring a substantial absolute increase over 
local methods ($+6.6\%$ for HMDB51, $+7.1\%$ for Holywood2, $+8.6\%$ for MPII 
Cooking, $+9.3\%$ for ChaLearn).\\


\noindent\textbf{Analysis of action classification results.}
\begin{figure}[t!]
    \centering
    \includegraphics[width=1.0\linewidth]{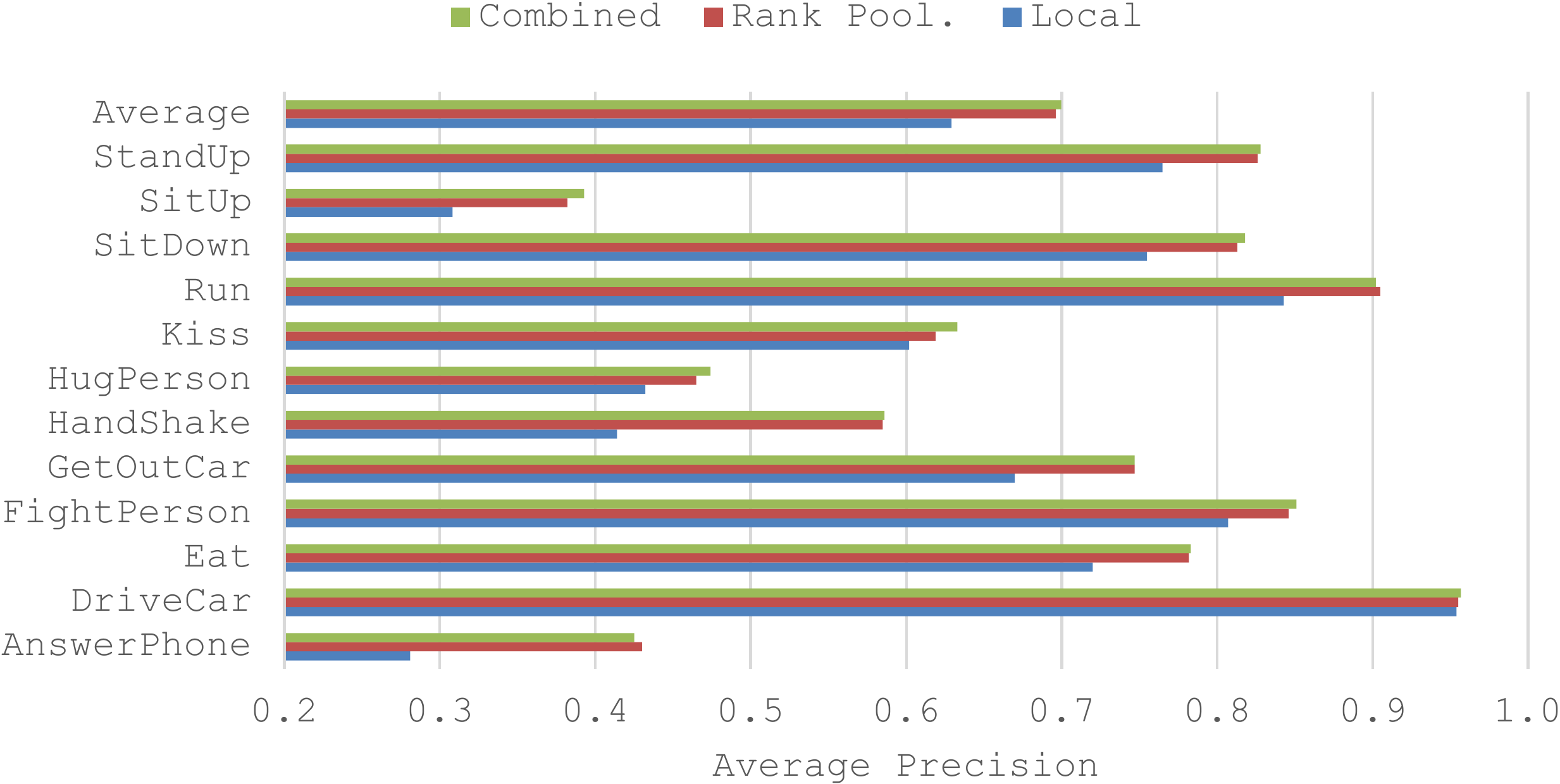}    
    \caption{Per class AP in the Hollywood2 dataset. The AP is improved significantly for all 
classes, with an exception of ``Drive car'', where context already provides useful information.}
    \label{fig:holywood2-class-accuracies}
\end{figure}
Looking at the individual results for the Hollywood2 dataset shown in Figure~\ref{fig:holywood2-class-accuracies}, we observe that almost all actions 
benefit the same, about a 7\% average increase. Some notable exceptions are 
 ``answer phone'', which improves by 14\% and ``handshake'', which
improves by 17\%. For ``drive car'' there is no improvement. The 
most probable cause is that the car context already provides enough evidence 
for the classification of the action, also reflected in the high classification 
accuracy of the particular action. Our method brings improvements for periodic 
actions such as ``run, handshake'' as well as non-periodic actions such as 
``get-out-of-car''.

\begin{table}[t]   
\scriptsize
\centering 
\begin{tabular}{l c c c c c c c}  \hline
& HOG	& HOF & MBH & TRJ & Combined \\ \hline  
\textit{Local}
& 39.2 & 48.7 & 50.8 & 36.0 & 55.2 \\
\textit{TP}
& 40.7 & 52.2 & 53.5 & 37.0 & 57.2 \\ \hline  
\textit{FDRP}
& 39.2 & 52.7 & 53.0 & 37.0 & 57.9 \\
\textit{RFDRP}
& 41.6 & 53.3 & 54.6 & 39.1 & 59.1 \\
\textit{NL-FDRP}
& 44.2 & 54.7 & 55.2 & 37.7 & 61.0 \\
\textit{NL-RFDRP}
& \textbf{46.6} & \textbf{55.7} & \textbf{56.7} & \textbf{39.5} & 
\textbf{61.6} \\ \hline
\textit{Local + FDRP}
& 42.4 & 53.7 & 54.3 & 39.7 & 59.3 \\
\textit{Local + RFDRP}
& 42.7 & 53.9 & 54.9 & 40.0 & 59.4 \\
\textit{Local + NL-FDRP}
& 45.6 & 56.2 & 56.2 & 41.0 & 61.3 \\
\textit{Local + NL-RFDRP}
& \textbf{47.0} & \textbf{56.6} & \textbf{57.1} & \textbf{41.3} & 
\textbf{61.8} \\ \hline
\end{tabular}  
\caption{One-vs-all accuracy on HMDB51 dataset~\cite{Kuehne2011}}
\label{tab:HMDB51}
\end{table}

\begin{table}[t]
\centering
\scriptsize
\begin{tabular}{l c c c c c c c} \hline
& HOG	& HOF & MBH & TRJ & Combined \\\hline     
\textit{Local}
& 47.8 & 59.2 & 61.5 & 51.2 & 62.9 \\
\textit{TP}
& 52.0 & 61.1 & 63.6 & 52.1 & 64.8 \\\hline  
\textit{FDRP}
& 45.3 & 59.8 & 60.5 & 49.8 & 63.6 \\
\textit{RFDRP}
& 50.5 & 63.6 & 65.5 & \textbf{55.1} & 67.9 \\
\textit{NL-FDRP}
& 52.8 & 60.8 & 62.9 & 50.2 & 65.6 \\
\textit{NL-RFDRP}
& \textbf{56.7} & \textbf{64.7} & \textbf{66.9} & 54.5 & \textbf{69.6} \\\hline 
\textit{Local + FDRP}
& 50.2 & 62.0 & 64.4 & 53.6 & 66.7 \\
\textit{Local + RFDRP}
& 52.7 & 64.3 & 66.2 & 55.9 & 68.7 \\
\textit{Local + NL-FDRP}
& 54.7 & 62.9 & 64.9 & 54.4 & 67.6 \\
\textit{Local + NL-RFDRP}
& \textbf{57.4} & \textbf{65.2} & \textbf{67.3} & \textbf{56.1} & 
\textbf{70.0} \\\hline
\end{tabular}   
\caption{Results in mAP on Hollywood2 dataset~\cite{marszalek09}}
\label{tab:holywood2}
\end{table}

\begin{table}[t]
\centering  
\scriptsize
\begin{tabular}{l c c c c c c c}\hline
& HOG	& HOF & MBH & TRJ & Combined \\\hline  
\textit{Local}
& 49.4 & 52.9 & 57.5 & 50.2 & 63.4 \\
\textit{TP}
& \textbf{55.2} & 56.5 & 61.6 & \textbf{54.6} & 64.8 \\\hline
\textit{FDRP}
& 50.7 & 53.5 & 58.0 & 48.8 & 62.4 \\
\textit{RFDRP}
& 53.1 & 55.2 & 61.4 & 51.9 & 63.5 \\
\textit{NL-FDRP}
& 52.8 & \textbf{60.8} & \textbf{62.9} & 50.2 & \textbf{65.6} \\
\textit{NL-RFDRP}
& 50.6 & 53.8 & 56.5 & 50.0 & 62.7 \\\hline
\textit{Local + FDRP}
& 61.4 & 65.6 & 69.0 & 62.7 & 71.5 \\
\textit{Local + RFDRP}
& \textbf{63.7} & \textbf{65.9} & \textbf{69.9} & \textbf{63.0} & 71.7 \\
\textit{Local + NL-FDRP}
& 63.5 & 65.0 & 68.6 & 61.0 & 71.8 \\
\textit{Local + NL-RFDRP}
& 64.6 & 65.7 & 68.9 & 61.2  & \textbf{72.0} \\ \hline
\end{tabular}  
\caption{Results in mAP on MPII Cooking fine grained action 
dataset~\cite{rohrbach2012database}. }
\label{tab:cooking}
\end{table}

\begin{table}[t]
\centering
\small
\begin{tabular}{l c c c} \hline
& Precision & Recall & F-score \\ \hline
\textit{Local}
& 65.9 & 66.0 & 65.9 \\
\textit{TP}
& 67.7 & 67.7 & 67.7 \\ \hline  
\textit{FDRP}
& 60.6 & 60.4 & 60.5 \\
\textit{RFDRP}
& 65.5 & 65.1 & 65.3 \\
\textit{NL-FDRP}
& 69.5 & 69.4 & 69.4 \\
\textit{NL-RFDRP}
& \textbf{74.0} & \textbf{73.8} & \textbf{73.9} \\ \hline

\textit{Local + FDRP}
& 71.4 & 71.5 & 71.4 \\
\textit{Local + RFDRP}
& 73.9 & 73.8 & 73.8 \\
\textit{Local + NL-FDRP}
& 71.8 & 71.9 & 71.8 \\
\textit{Local + NL-RFDRP}
& \textbf{75.3} & \textbf{75.1} & \textbf{75.2} \\ \hline
\end{tabular}
\caption{Detailed analysis of precision and recall on the ChaLearn gesture 
recognition dataset~\cite{Escalera2013}}
\label{tab:gesture}
\end{table}

\begin{figure}[t!]
    \centering
    \includegraphics[width=0.49\linewidth]{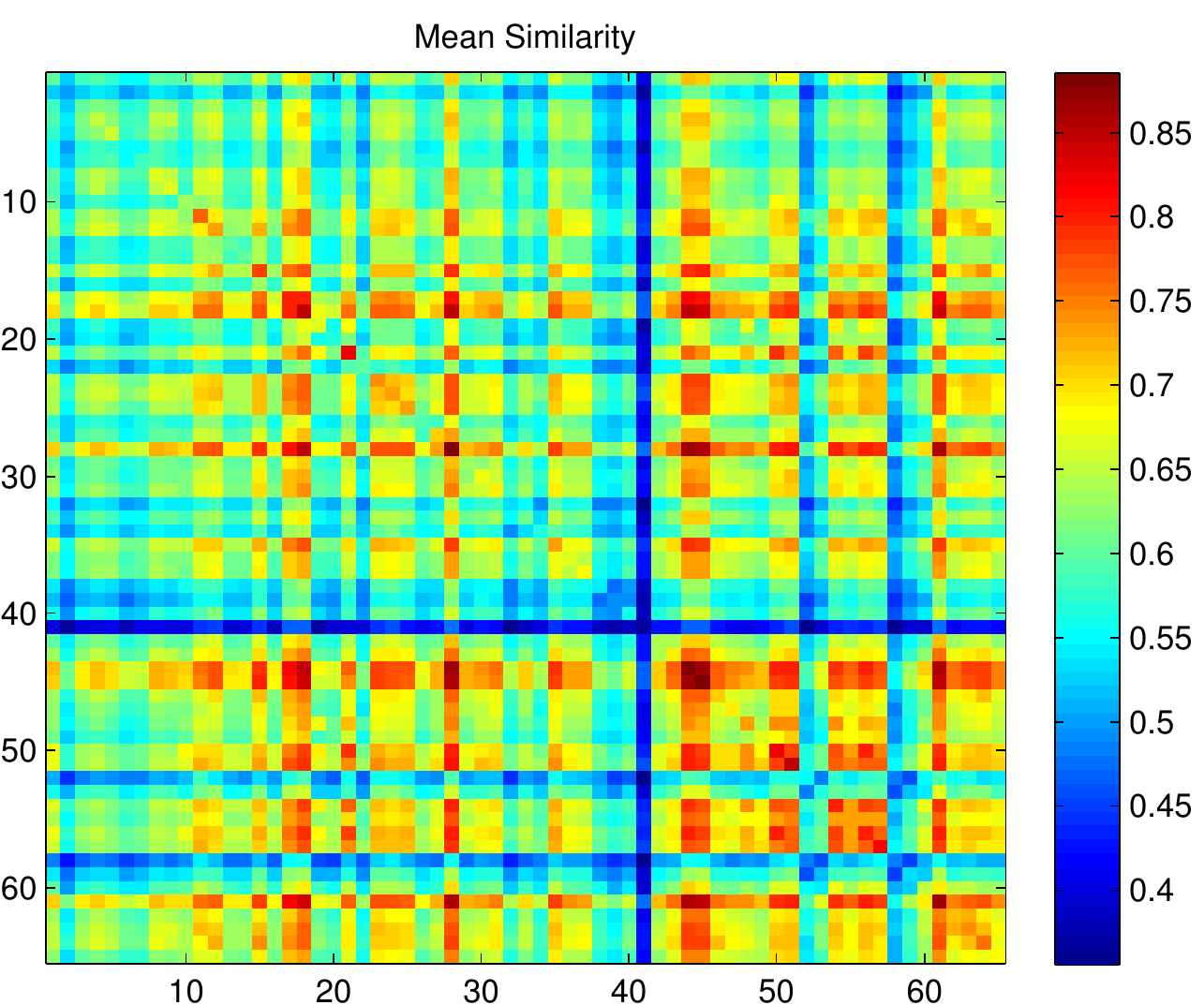}
    \includegraphics[width=0.49\linewidth]{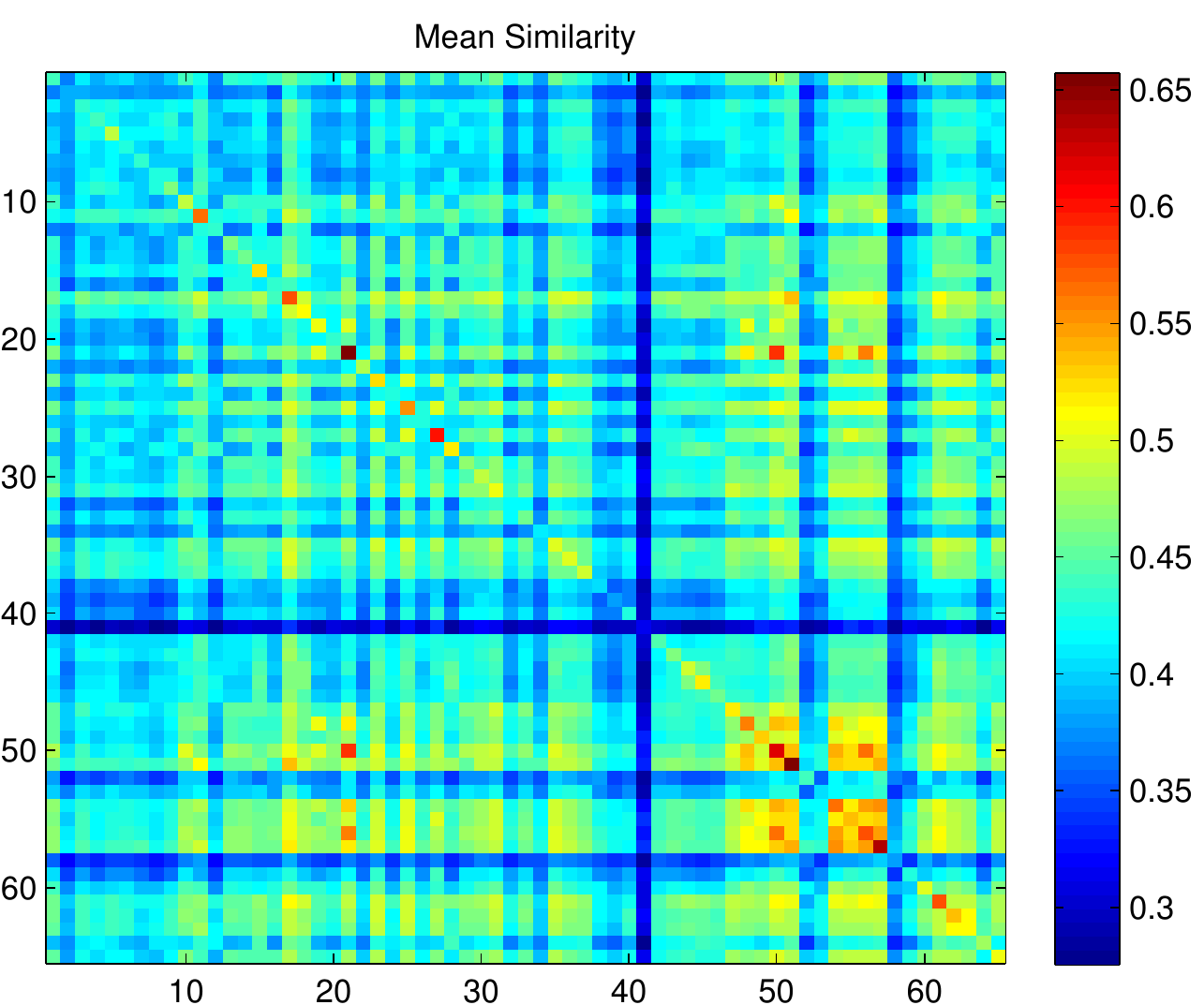}        
    \caption{Mean class similarity obtained with (left) max-pooling and (right) 
\name on MPII Cooking activities dataset using BOW-based MBH features extracted 
on dense trajectories. Non-linear forward \name are used for our method.}
    \label{fig:cooking-mcs}
\end{figure}

For the case of the ChaLearn dataset (Table \ref{tab:gesturesoa}), we see that 
\name is able to achieve superior results without requiring to explicitly define
task-specific steps, e.g. hand-posture or hand-trajectory modeling \cite{Martinez2015}.

To gain further insight we investigate the mean similarity 
computed over classes on MPII cooking dataset with BOW-based MBH features. 
We construct the dot product kernel matrix using all the samples and 
then compute the mean similarity between classes, see 
Figure~\ref{fig:cooking-mcs}. The \name kernel matrix 
(Figure~\ref{fig:cooking-mcs} (right)) appears to be more discriminative than 
the one with max-pooled features (Figure~\ref{fig:cooking-mcs} (left)). The 
action ``smell'' (\#41) seems very difficult to discriminate 
either using max-pooling or \name method. Actions ``sneeze'' (\#44) and 
``stamp'' (\#45) seem to be very similar in-terms of appearances, however with 
\name we can discriminate them better. Actions like ``take \& put in 
cupboard'' (\#47), ``take \& put in drawer''(\#48), ``take \& put in fridge'' 
(\#49) and ``take \& put in oven'' (\#50) seem to be the most confused ones for 
\name. These actions differ in the final instrument, but not in the dynamics of 
the action. 

\subsection{\Name analysis}
\label{sec.action.exp.stability}
\noindent\textbf{Stability to dropped frames} 
We analyze the stability of \name compared to average 
pooling and temporal pyramids. For this experiment we use 
Hollywood2 dataset and MBH features with Fisher vectors. We gradually 
remove 5\%, 10\%, $\ldots$ 25\% of random frames from each video from both train and test sets  and then 
measure the change in mean average precision. 

\begin{figure}[t!]
    \centering
    \includegraphics[width=0.8\linewidth]{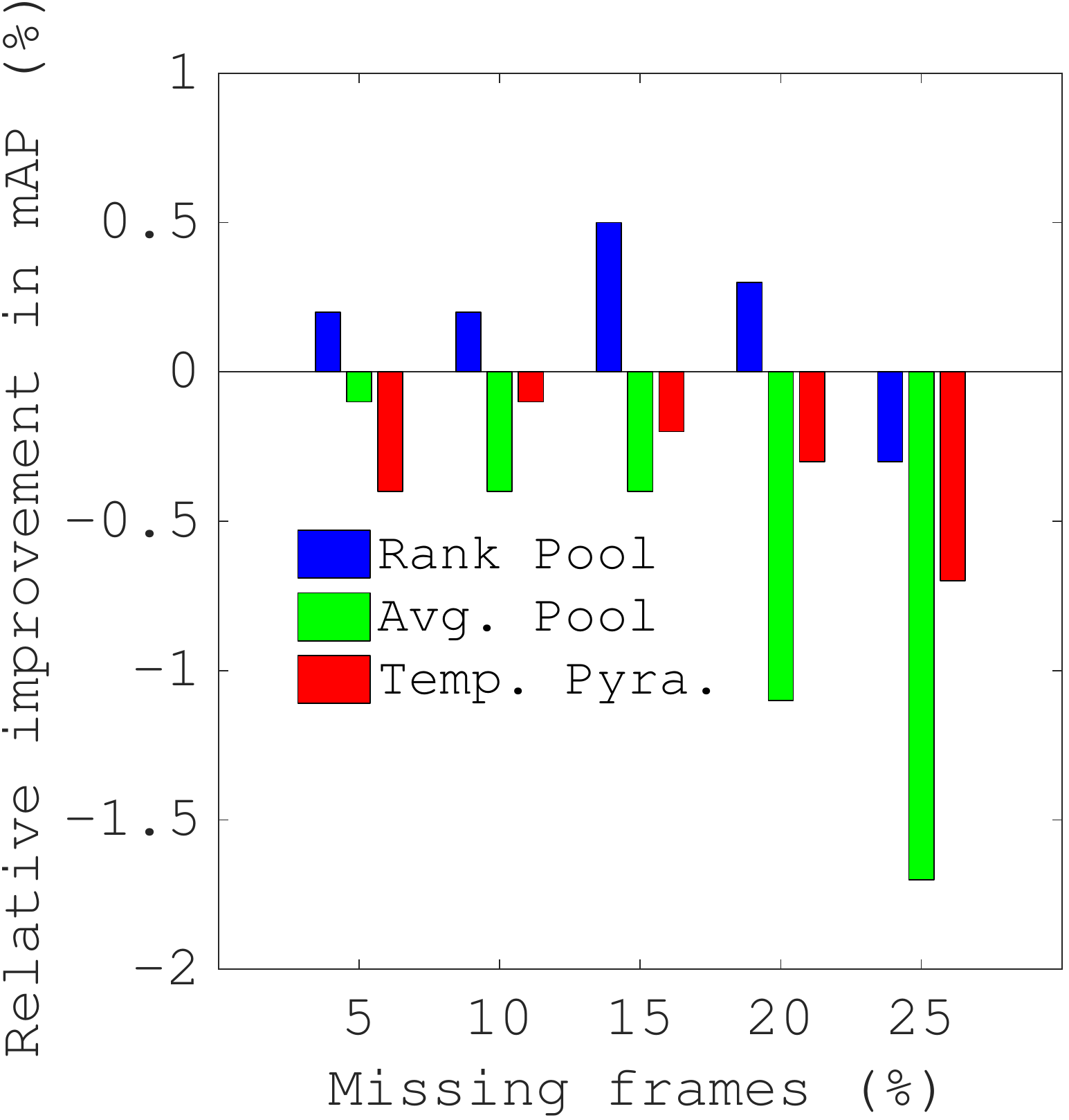}    
    \caption{Comparison of action recognition performance after removing some 
frames from each video randomly on Hollywood2. \name appears to be stable even 
when up to 20\% of the frames are missing.}
    \label{fig:stability}
\end{figure}

We present in Figure~\ref{fig:stability} the relative change in mAP after 
frame removal. Typically, we would expect the mAP to decrease. 
Interestingly, removing up-to 20\% of the frames from the video 
does not significantly change the results of \name; in-fact we observe a 
slight relative improvement. This is a clear indication of the stability of \name 
and an advantage of learning-based temporal pooling. As expected, the mAP 
decreases for both average pooling method and the temporal pyramids method as 
the number of frames that are removed from videos increases. For average 
pooling mAP seems to drop almost in an exponential manner. However, it should 
be noted that 25\% of the video frames is a significant amount of data. We 
believe the results illustrate the stability of \name.\\

\begin{figure}[t!]
    \centering
    \subfloat[]{\includegraphics[width=0.75\linewidth]{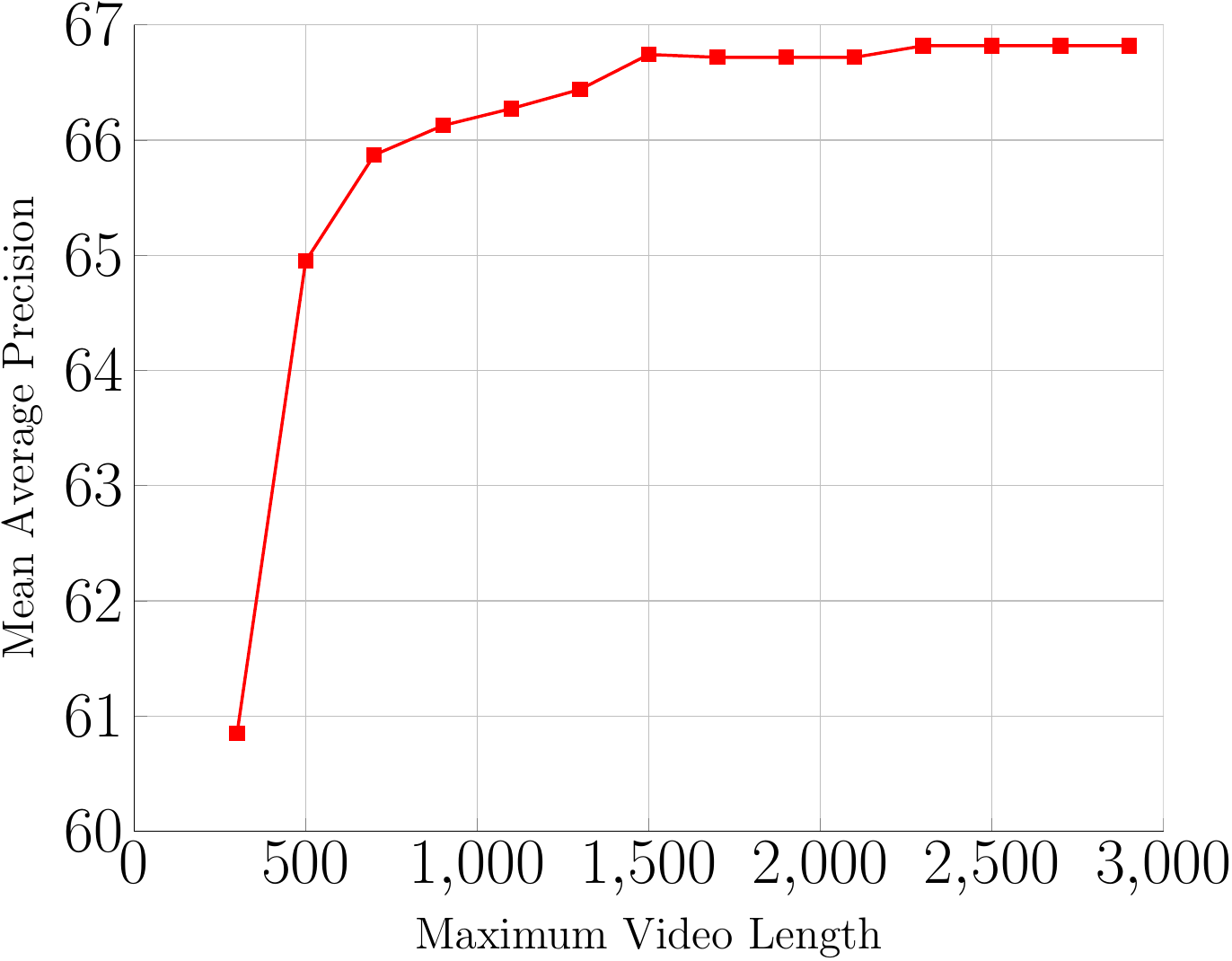}} 
    \caption{Hollywood2 action recognition performance with respect to the length of the video using our \name method. }
    \label{fig:max-len-vs-map}
\end{figure}

\noindent\textbf{Effect of video length.} 
In this experiment we analyse how the length of the video influences the testing performance. We train \name-based classifiers as before using the entire training set and then partition the test set into three segments. Then, we compare the action classification accuracies obtained with different video lengths. Results are shown in Figure~\ref{fig:max-len-vs-map}. Interestingly, the longer the video, the better our method seems to perform. This is not as surprising, since longer videos are more likely to contain more dynamic information compared to shorter videos. Also, for longer videos averaging will likely be more affected by outliers. What is more noteworthy is the relative difference in accuracy between very long and very short videos, approximately 6\%. We conclude that our method is capable of capturing the dynamics of short videos as well as of long videos.   \\

\noindent\textbf{The impact of feature maps on Fisher Vectors.}
In this section we evaluate the effect of different feature maps during ranker function construction and final video classification. We use MBH features as the representation and evaluate the activity recognition performance on Hollywood2 dataset. Results are reported in Table~\ref{tab:feat.maps}.

We observe that the combination of posneg feature maps both for computing \name, as well as computing the final classification kernel, outperforms all other alternatives. The closest competitor is when we use the posneg kernel for computing the \name features. In general, we observe that for the classification kernel the different combinations perform somewhat similarly, given a fixed \name feature map. We conclude the highest accuracies with \name are obtained when we apply the posneg feature map, irrespective to the classification kernel.\\

\begin{table}[t]
\centering
\small
\begin{tabular}{ c c c} \hline
Ranking Feature Map & Classifier Feature Map & mAP \\ \hline
$\sqrt{|x|}$ & $\text{sgn}(u) \sqrt{|u|}$ & 50.0 \\
$\sqrt{|x|}$ & $\sqrt{|u|}$ & 54.0 \\
$\sqrt{x^*}$ & $\sqrt{u^*}$ & \textbf{66.1} \\
$\sqrt{x^*}$ & $\text{sgn}(u) \sqrt{|u|}$ & 65.4  \\
$\sqrt{x^*}$ & $\sqrt{|u|}$ & 63.7 \\
\hline
\end{tabular}
\caption{Comparison of different features maps for ranking and classification. We use different symbols, $x$ and $u$, to avoid the confusion, as $u$ refers to the feature encodings (\eg, Fisher vectors) that we use to compute \name, while $u$ refers to the \name features. $x^*$ stands for the the input to the posneg kernel, namely $x^*=[x^+, x^-]^T$.} 
  \label{tab:feat.maps}
\end{table}


\begin{table}[t]
\centering
\small
\begin{tabular}{ l c } \hline
Parameter pooling  	& mAP \\ \hline
Rank Pooling with RankSVM  & 66.0 \\
Rank Pooling with SVR	& 66.5 \\ 
Subspace pooling 	& 56.4 \\
Robust Subspace Pooling & 64.1 \\
HMM pooling		& 17.8 \\
Neural Network pooling 	& 21.1\\ \hline

\end{tabular}
\caption{Pooling parameters as representations from different parametric models. R-PCA stands for the Robust PCA.
Experiments were conducted on the Hollywood2 dataset using MBH features.} 
  \label{tab:other.baseline}
\end{table}

\noindent\textbf{Functional parameters as temporal pooling.}
\label{sec.action.exp.other.baseline}
In this experiment we evaluate several parametric models in which we can use the parameters to represent a video. More specifically, we evaluate rank pooling using SVR~\cite{Smola2004} and RankSVM~\cite{JoachimsKDD2006} subspace pooling using the first principal eigenvector only as the video representation~(Sec.~\ref{sec:parameter-pooling}) and robust subspace pooling using the first eigenvector only as described in equation.~\ref{eq:pca2}. Additionally, we use the parameters of two layered fully connected neural networks as a video representation. In this case, the neural network consists of one hidden layer (10 hidden units) and the input layer. It is trained to map frame data to the time variable hoping to capture dynamics similar to SVR~\cite{Smola2004}. Furthermore, we train a Hidden Markov Model using the input video data and then use the transition and observation probability matrix as a video representation. We run the experiment on Hollywood2 dataset with MBH features and show results in Table~\ref{tab:other.baseline}.

We observe that standard PCA-based subspace pooling is less accurate than both the SVR and the RankSVM rank pooling.
The robust subspace pooling, which deals better with very low data volume to dimensionality ratios, captures the video-wide temporal evolution reasonably well.
However, pooling from ranker SVR machines works best.
Interestingly, the neural network and HMM performance is poor. Probably, the neural network overfits easily compared to the SVR machines.

We conclude that for moderately long videos using the parameters of simpler, linear machines as the representation for the sequence data is to be preferred to avoid overfitting. However, we expect that for very long videos or for even richer frame representations more complex dynamics could arise. In these cases higher capacity methods, like neural networks, would likely capture better the underlying dynamics.\\


\noindent\textbf{CNN features for action classification}
\label{sec.action.exp.cnn}
In this experiment we evaluate our method using the convolutional 
neural network~(CNN)-based features. We use the activations of the first 
fully connected layer of vgg-16 network~\cite{Simonyan2014a} to 
represent each frame in a video. We compare several pooling techniques 
using Hollywood2 dataset in Table~\ref{tab:cnn}. \Name by itself does 
not perform that well compared to local (average pooling) method 
(32.2 mAP vs. 39.0 mAP). However, the combination of \name with the local 
approach improves the results to 46.4 mAP. The CNN features used in 
this experiment are 4096 dimensional and are not fine tuned for 
action classification. 
As the pre-trained features are trained specifically for 
appearance-based classification, we combine CNN features 
with MBH features. With the local 
approach, the combination of CNN and MBH results in 65.6 mAP. 
The best results are obtained with local pooling of CNN and temporal 
pooling of MBH. We believe this strategy exploits the advantage 
of both appearance information and dynamics of videos.\\

\begin{table}[t]
\centering
\small
\begin{tabular}{ l c } \hline
Method & mAP \\ \hline
Local(cnn) & 39.0  \\ 
NL-RFDRP(cnn) & 32.2 \\ 
Local(cnn)+NL-RFDRP(cnn) & 46.4  \\ 

Local(cnn)+Local(MBH) & 65.6  \\ 
Local(cnn)+NL-RFDRP(MBH) & \textbf{70.1}  \\ 
Local(cnn+MBH)+NL-RFDRP(MBH) & 69.7  \\ 
Local(cnn+MBH)+NL-RFDRP(cnn+MBH) & 69.5  \\ \hline

\end{tabular}
\caption{Results obtained on Hollywood2 dataset using CNN (vgg-16 network~\cite{Simonyan2014a}) features. } 
  \label{tab:cnn}
\end{table}

\subsection{State-of-the-art and discussion.}
\label{sec.action.exp.soa}
Last, we compare the nonlinear forward and reverse \name combined with the 
local temporal information with the latest state-of-the-art in action 
recognition. We summarize the results in Table~\ref{tab:soa} and 
Table~\ref{tab:gesturesoa}. Note that for Hollywood2 and HMDB51, we use data augmentation by mirroring the videos as in~\cite{hoi2014}, which brings a further 5\% improvement, and combine with max-pooled CNN features to capture static appearance information explicitly.

\begin{table}[t]
\centering
\small
\begin{tabular}{l c c c} \hline
& HMDB51 & Hollywood2 & Cooking \\ \hline
\textit{\Name}+CNN & 65.8 & \textbf{75.2} & -- \\
\textit{\Name} & 63.7 & 73.7 & \textbf{72.0} \\
\textit{Hoai \etal~\cite{hoi2014}} & 60.8 & 73.6 & -- \\
\textit{Peng \etal~\cite{PengECCV2014} }
& \textbf{66.8} & -- & -- \\
\textit{Wu \etal~\cite{Wu_2014_CVPR} }
& 56.4 & -- & -- \\
\textit{Jain \etal~\cite{jain2013better}}
& 52.1 & 62.5 & -- \\
\textit{Wang \etal~\cite{wang2013action} }
& 57.2 & 64.3 & -- \\
\textit{Wang \etal~\cite{wang2013dense}}
& 46.6 & 58.2 & -- \\
\textit{Taylor \etal~\cite{taylor2010convolutional}}
& -- & 46.6 & -- \\
\textit{Zhou \etal~\cite{Zhou14eccv} }
& -- & -- & 70.5 \\
\textit{Rohrbach \etal~\cite{rohrbach2012database}}
& -- & -- & 59.2 \\\hline
\end{tabular}
\caption{Comparison of the proposed approach with the state-of-the-art 
methods sorted by reverse chronological order. Results reported in mAP for Hollywood2 and Cooking datasets. For HMDB51 we report one-vs-all classification accuracy.} 
  \label{tab:soa}
\end{table}

\begin{table}[th!]
\small
\centering 
\begin{tabular}{l c c c}\hline
& Precision & Recall & F-score \\\hline  
  \textit{\Name}
  & \textbf{75.3} & \textbf{75.1} & \textbf{75.2} \\
  \textit{Martinez-Camarena \etal~\cite{Martinez2015}}
  & 61.4 & 61.9 & 61.6 \\
  \textit{Pfister \etal~\cite{pfister2014domain}}
  & 61.2 & 62.3 & 61.7 \\
  \textit{Yao \etal~\cite{Yao14cvpr}}
  & -- & -- & 56.0 \\
  \textit{Wu \etal~\cite{Wu:2013:FMF}}
  & 59.9 & 59.3 & 59.6 \\  \hline  
  \end{tabular}  
   \caption{Comparison of the proposed approach with the state-of-the-art 
methods on ChaLearn gesture recognition dataset sorted by reverse chronological 
order.}
  \label{tab:gesturesoa}
\end{table}

By the inspection of Tables~\ref{tab:soa} and \ref{tab:gesturesoa}, as well as from the results in the 
previous experiments, we draw several conclusions. First, \name is useful and robust 
for encoding video-wide, temporal information. Second, \name is complementary to action recognition methods that compute local temporal 
features, such as improved trajectory-based features~\cite{wang2013action}. 
In fact, fusing \name with the previous state-of-the-art in local motion and 
appearance, we improve up to 10\%. Third, \name is complimentary with static feature representations such as CNN-based max pooled features. Forth, \name is only outperformed on HMDB51 by~\cite{PengECCV2014}, who combine their second layer Fisher vector features 
with normal Fisher vectors to arrive at 205K dimensional vectors and a 66.8\% 
accuracy. When using Fisher vectors like \name does, Peng \etal.~\cite{PengECCV2014} 
obtain 56.2\%, which is 10\% lower than what we obtain with \name.

%
%
%
%


\section{Discussion and Conclusion}
\label{sec:action-conclusion}
We introduce \name, a new pooling methodology that models the evolution of appearance and dynamics in a video. Rank pooling is an unsupervised, learning based temporal pooling method, which aggregates the relevant information throughout a video via fitting learning-to-rank models and using their parameters as the new representation of the video. We show the regularized learning of the learning-to-rank algorithms, as well as the minimization of the temporal ordering empirical risk, has in fact favorable generalization properties that allow us to capture robust temporal and video dynamics representations. Moreover, we show that the ranking models can be replaced with different parameteric models, such as principal component analysis. However, experiments reveal that learning-to-rank linear machines seem to capture the temporal dynamics in videos best. We demonstrate that a temporal smoothing and sequence pre-processing is important for modelling the temporal evolution in sequences. Last, we show that designing kernels that separate the positive from the negative part of the incoming features has a substantial effect on the final classification using \name. Based on extensive experimental evaluations on different datasets and features we conclude that, our method is applicable to a wide variety of frame-based representations for capturing the global temporal information of a video.

%

In the current work we focused mainly on exploring \name within an action classification setting on moderately long videos. However, we believe that \name could easily be exploited in other tasks too, such as video caption generation, action detection, video retrieval, dynamic texture and video summarization.

We conclude that \name is a novel and accurate method for capturing the temporal evolution of appearances and dynamics in videos.


\section{Acknowledgment}
The authors acknowledge the support of the Australian Research Council Centre of Excellence for Robotic Vision (project number CE140100016), FP7 ERC Starting Grant 240530 COGNIMUND, KU Leuven DBOF PhD fellowship, the FWO project \textit{Monitoring of abnormal activity with camera systems} and iMinds High-Tech Visualization project.



\end{document}